\documentclass[letterpaper]{article} 
\usepackage{aaai25}  
\usepackage{times}  
\usepackage{helvet}  
\usepackage{courier}  
\usepackage[hyphens]{url}  
\usepackage{graphicx} 
\usepackage{natbib}  
\usepackage{caption} 
\usepackage{algorithm}
\usepackage{algorithmic}

\usepackage{newfloat}
\usepackage{listings}
\DeclareCaptionStyle{ruled}{labelfont=normalfont,labelsep=colon,strut=off} 
\floatstyle{ruled}
\newfloat{listing}{tb}{lst}{}
\floatname{listing}{Listing}
\usepackage{amsmath}
\usepackage{amsfonts}
\usepackage{amssymb}
\usepackage{tabularx}
\usepackage{tabularray}
\usepackage{xcolor}
\usepackage{xspace}
\usepackage{booktabs}
\usepackage{makecell}
\usepackage{animate}
\newcommand{\ie}{i.e.\xspace}
\newcommand{\eg}{e.g.\xspace}
\newcommand{\methodName}{CAGE\xspace}
\newcommand{\methodNameExt}{visual Composition and Animation for video GEneration\xspace}

\title{CAGE: Unsupervised Visual Composition and Animation\\ for Controllable Video Generation}
\author {
    Aram Davtyan\textsuperscript{\rm 1},
    Sepehr Sameni\textsuperscript{\rm 1},
    Bj\"orn Ommer\textsuperscript{\rm 2},
    Paolo Favaro\textsuperscript{\rm 1}
}
\affiliations {
    \textsuperscript{\rm 1}Computer Vision Group, Institute of Informatics, University of Bern, Switzerland\\
    \textsuperscript{\rm 2}CompVis @ LMU Munich and MCML, Germany\\
    aram.davtyan@unibe.ch, sepehr.sameni@unibe.ch, b.ommer@lmu.de, paolo.favaro@unibe.ch
}

\begin{document}

\maketitle

\begin{abstract}
The field of video generation has expanded significantly in recent years, with controllable and compositional video generation garnering considerable interest. Most methods rely on leveraging annotations such as text, objects' bounding boxes, and motion cues, which require substantial human effort and thus limit their scalability. In contrast, we address the challenge of controllable and compositional video generation without any annotations by introducing a novel unsupervised approach. Our model is trained from scratch on a dataset of unannotated videos. At inference time, it can compose plausible novel scenes and animate objects by placing object parts at the desired locations in space and time. The core innovation of our method lies in the unified control format and the training process, where video generation is conditioned on a randomly selected subset of pre-trained self-supervised local features. This conditioning compels the model to learn how to inpaint the missing information in the video both spatially and temporally, thereby learning the inherent compositionality of a scene and the dynamics of moving objects. The abstraction level and the imposed invariance of the conditioning input to minor visual perturbations enable control over object motion by simply using the same features at all the desired future locations. We call our model CAGE, which stands for visual Composition and Animation for video GEneration. We conduct extensive experiments to validate the effectiveness of CAGE across various scenarios, demonstrating its capability to accurately follow the control and to generate high-quality videos that exhibit coherent scene composition and realistic animation.
\end{abstract}

\begin{links}
\link{Project website}{https://araachie.github.io/cage}
\end{links}

\begin{figure}[t]
\centering
\footnotesize
\newcommand\curWidth{0.14\linewidth}
\begin{tabular}{@{}c@{\hspace{0.3mm}}c@{\hspace{0.3mm}}c@{\hspace{0.3mm}}c@{\hspace{0.3mm}}c@{\hspace{0.3mm}}c@{\hspace{0.3mm}}c@{}}
    \makecell{1st\\ source} & \makecell{2nd\\ source} & \makecell{3rd\\ source} & control & \multicolumn{3}{c}{generated sequence $\rightarrow$} \\
     \includegraphics[width=\curWidth]{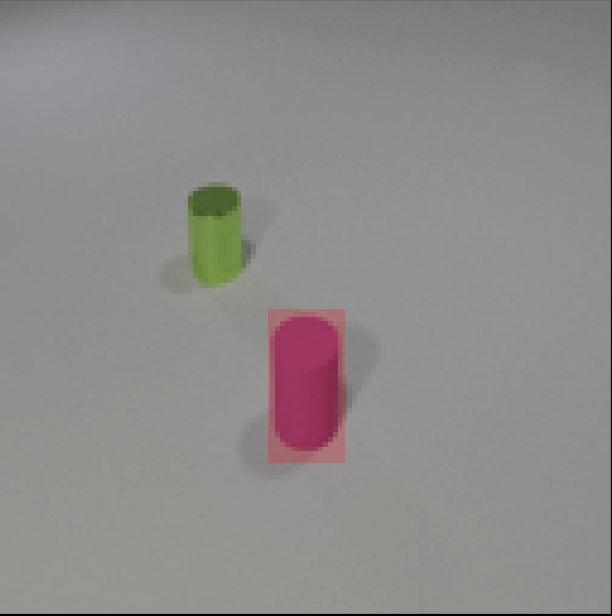} &
     \includegraphics[width=\curWidth]{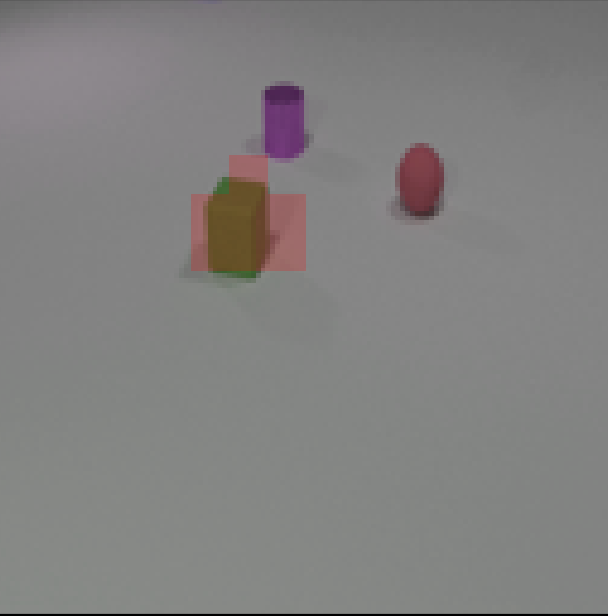} & 
     \includegraphics[width=\curWidth]{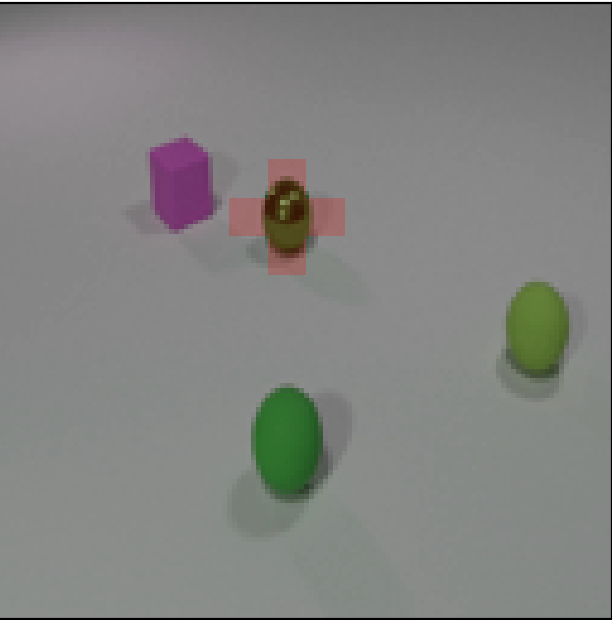} &
     \includegraphics[width=\curWidth]{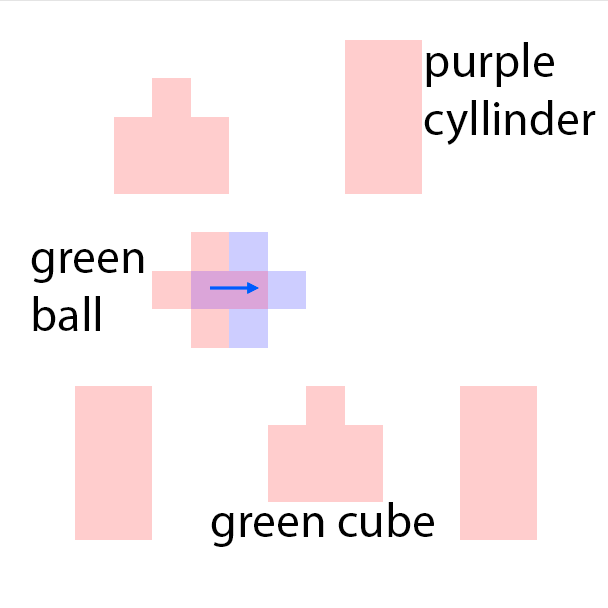} & 
     
     \animategraphics[width=\curWidth]{7}{Figures/cl_comp_3_obj/image_0000}{0}{8} & 
     \includegraphics[width=\curWidth]{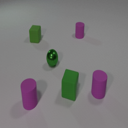} & 
     \includegraphics[width=\curWidth]{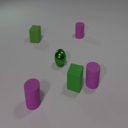} \\

     

     \includegraphics[width=\curWidth]{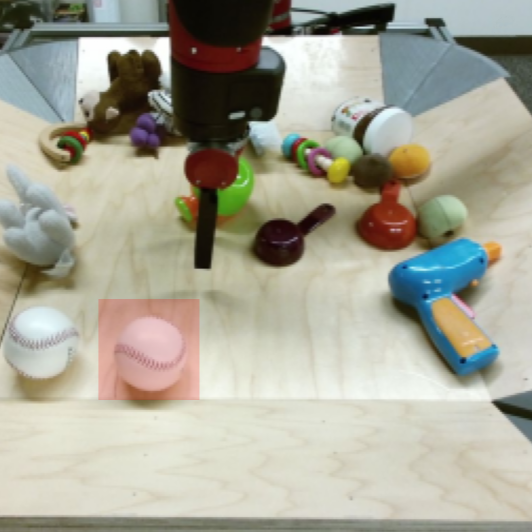} &
     \includegraphics[width=\curWidth]{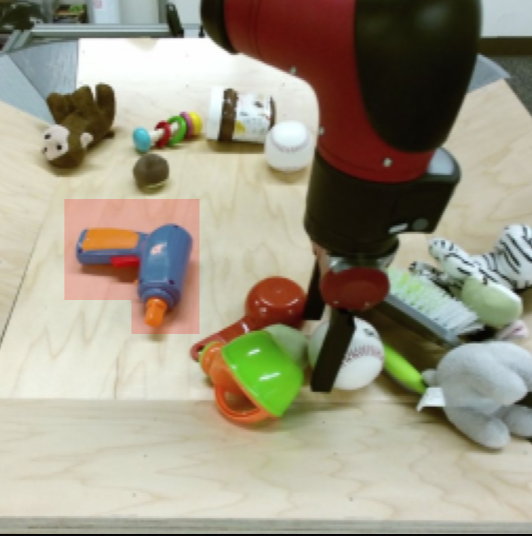} & 
     \includegraphics[width=\curWidth]{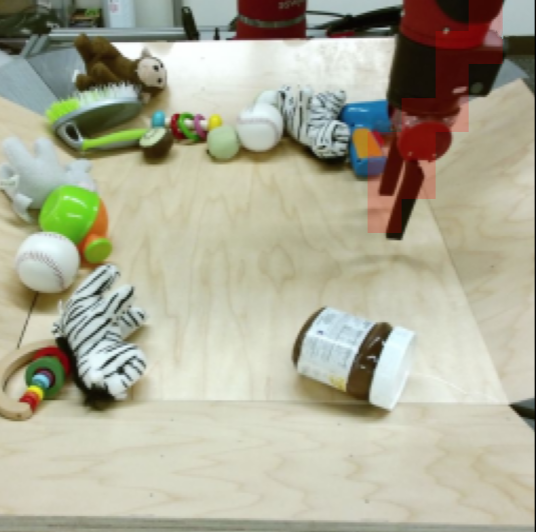} &
     \includegraphics[width=\curWidth]{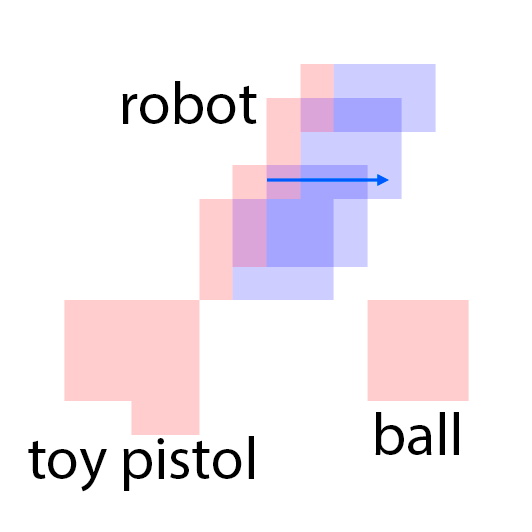} & 
     
     \animategraphics[width=\curWidth]{7}{Figures/bair_comp_bpr/image_0000}{0}{8} & 
     \includegraphics[width=\curWidth]{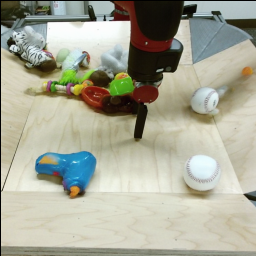} & 
     \includegraphics[width=\curWidth]{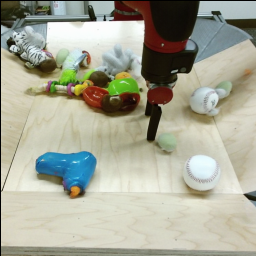} \\

     
\end{tabular}
\caption{Scene composition and animation with \methodName on the CLEVRER and the BAIR datasets. \methodName is able to combine multiple object features from different source images and use them to compose and animate the scene in a controllable way. The selected features are shown as overlaying red patches. Blue patches in the controls correspond to the intended future locations of the objects. Notice the ability of the model to carefully adjust the appearances (\eg, sizes, shadows and lights) of the objects based on their location in the target layout. Use the Acrobat Reader to play the  first images in the generated sequences as videos.}
\label{fig:comp}
\end{figure}

\section{Introduction}
\label{sec:intro}


Video generation has gained significant attention in recent years, offering a transformative approach to various domains, ranging from content creation \cite{bar2024lumiere} to robotics \cite{GuoTAMP2023}, autonomous driving\footnote{For example, GAIA by Wayve (\url{https://wayve.ai}) and the world models developed by Waabi (\url{https://waabi.ai})} \cite{hu2023gaia1,zhang2024learning,gao2024vista} and video games \cite{menapace2024promptable}.
Controllable video generation models are of particular interest, as they enable users to simulate the outcomes of desired changes in the environment, such as editing scene composition or assigning specific actions (\eg, moving the agent or other objects in a certain direction). Unlike traditional simulators, which require extensive manual effort to design and maintain synthetic environments, controllable video generation models can learn directly from real-world data, capturing interactions within complex scenes. This capability allows for the creation of realistic and dynamic environments that can be used to train and evaluate intelligent agents in a more flexible manner~\cite{ha2018worldmodels, mendonca2023structured}.

However, integrating control presents challenges due to the unavailability of information about ongoing actions in real-world video data. Existing models often rely heavily on large-scale and expensive supervision, such as text-annotations~\cite{Hu2021MakeIM}, objects' bounding boxes~\cite{Wang2024BoximatorGR,li2024animate}, segmentation~\cite{han2022show}, and motion cues~\cite{Shi2024MotionI2VCA}.
These requirements not only limit the scalability and flexibility of these models but also constrain their applicability to new domains where such annotated data is scarce or nonexistent.

To address this limitation, recent advancements have proposed unsupervised learning methods that build controllable video generation models only from real videos (\ie, without any information about the actions or the action space) \cite{menapace2021playable,Blattmann2021iPOKEPA,davtyan2022glass,Bruce2024GenieGI}. In these models, control is often defined as a separate input from visual data, ranging from motion encodings \cite{Blattmann2021iPOKEPA,davtyan2022glass} to general embeddings \cite{menapace2021playable,menapace2022playable,Bruce2024GenieGI}, which are learned directly from the visual data.
However, these control signal choices impose limitations on the tasks the model can perform. For instance, they allow to specify objects' motion, but not how to compose a scene, which we aim to address in this work.

We propose \methodName, short for \textit{\methodNameExt}, a generative model capable of creating environments and animating objects within them (see Fig.~\ref{fig:comp}). A key innovation in \methodName is that its control is specified directly through a set of ``visual tokens'' rather than embeddings from a separate action space.  
These visual tokens provide information regarding the identity of objects (\emph{what}), or parts thereof, and their spatio-temporal placement provides information about \emph{where} the objects should appear in certain frames in the future.
As visual tokens we use DINOv2 spatial features \cite{Oquab2023DINOv2LR}. Our experiments demonstrate that leveraging DINOv2 features enables our model to mitigate overfitting to the control signal and facilitates zero-shot transfer from other image domains. During training, we extract DINOv2 features from future video frames and utilize a sparse subset of these features as the control signal. By using this partial information, the model is trained to inpaint the surrounding context, both spatially and temporally, and thus learns to generate future frames consistent with the specified control signal. In summary, \methodName produces a video output that encapsulates the desired composite scene, along with animations depicting the objects within it, including their interactions.
Because we train \methodName in a fully unsupervised manner, it can be readily scaled to accommodate large datasets.
Our main contributions are summarized as follows:
\begin{itemize}
    \item \methodName is a novel controllable video generation model trained without any human supervision, capable of scene composition and  object animation;
    \item we introduce a unified control format that simultaneously can specify how to compose and animate a scene through a sparse set of visual tokens. This format allows to describe a wider range of prediction tasks than motion-based controls. In particular, it allows to compose scenes. Moreover, our control allows zero-shot transfer from frames not in our training set;
    \item  \methodName generates more realistic videos than prior work (according to the commonly used metrics) and its controllability has been validated experimentally.
\end{itemize}

\section{Prior Work}
\label{sec:prior_work}
\noindent\textbf{Video Generation.} Recent advancements in the field of video generation have been remarkable, driven predominantly by the impressive capabilities of diffusion and flow-based models. These models have significantly enhanced the generation of both images and videos from text descriptions, setting a new benchmark in the domain. The foundational work in text-to-image generation~\cite{rombach2022high} has paved the way for its extension to the text-to-video arena~\cite{guo2023animatediff,ho2022imagen,ho2022video}. 
However, as highlighted by Blattmann \textit{et al.}~\cite{blattmann2023stable} and Zhang \textit{et al.}~\cite{zhang2023show}, despite these advancements, current text-to-video models often produce videos with objects whose motion trajectories appear relatively random. Moreover, specifying precise motion dynamics with a lone text prompt poses significant challenges. 

\noindent\textbf{Supervised Controllable Video Generation.} 
A first approach that can exploit more accurate motion specifications in the text prompt is MAGE~\cite{Hu2021MakeIM}. This approach aligns well with the text-to-video generation pipeline, albeit necessitating increased supervision. An alternative strategy 
is to use 
ControlNet~\cite{Zhang2023AddingCC} for videos~\cite{Zhang2023ControlVideoTC,Chen2023ControlAVideoCT,Ma2023MagicStickCV}. This is particularly effective when the future frame structure is known, as it offers a predefined control mechanism. 
Models such as Boximator~\cite{Wang2024BoximatorGR} and Motion-I2V~\cite{Shi2024MotionI2VCA} enhance controllability by employing bounding boxes or motion flows as more intuitive control mechanisms. 
These models 
rely on supervised learning paradigms and in particular on text descriptions. While effective, this reliance poses inherent limitations on scalability and adaptability, as supervised methods require large, labeled datasets, which are resource-intensive to produce and may reduce their applicability scope. 

\noindent\textbf{Unsupervised Controllable Video Generation.} Unsupervised approaches in video generation have been gaining traction, primarily focusing on leveraging implicit signals such as masks~\cite{huang2022layered}, learned actions~\cite{menapace2021playable,Bruce2024GenieGI, davtyan2022glass} and optical flow~\cite{Blattmann2021iPOKEPA}. These methods offer a promising avenue by minimizing reliance on extensive labeled datasets, thus addressing the scalability and adaptability challenges inherent in supervised models. 
The closest to our work, YODA~\cite{davtyan2024learn} conditions the generation on a sparse set of optical flow vectors and achieves remarkable capabilities in generating out-of-distribution and counterfactual scenarios.
However, despite the steep progress, these models often struggle to achieve cross-scene generalization, as their generative capabilities are tightly coupled with the specificity of the conditioning signals used during training. Moreover, none of these models is capable of simultaneously composing and animating the scene with the same control signal. In contrast, our model leverages sparse “visual tokens” for local conditioning, offering an innovative, intuitive and flexible mechanism that expands the control space available in the prior work and hence advances video generation. 

\section{Inside the \methodName}
\label{sec:method}

\begin{figure*}[t]
    \centering
    \includegraphics[width=0.67\linewidth, trim=3cm 0cm 3cm 0cm, clip]{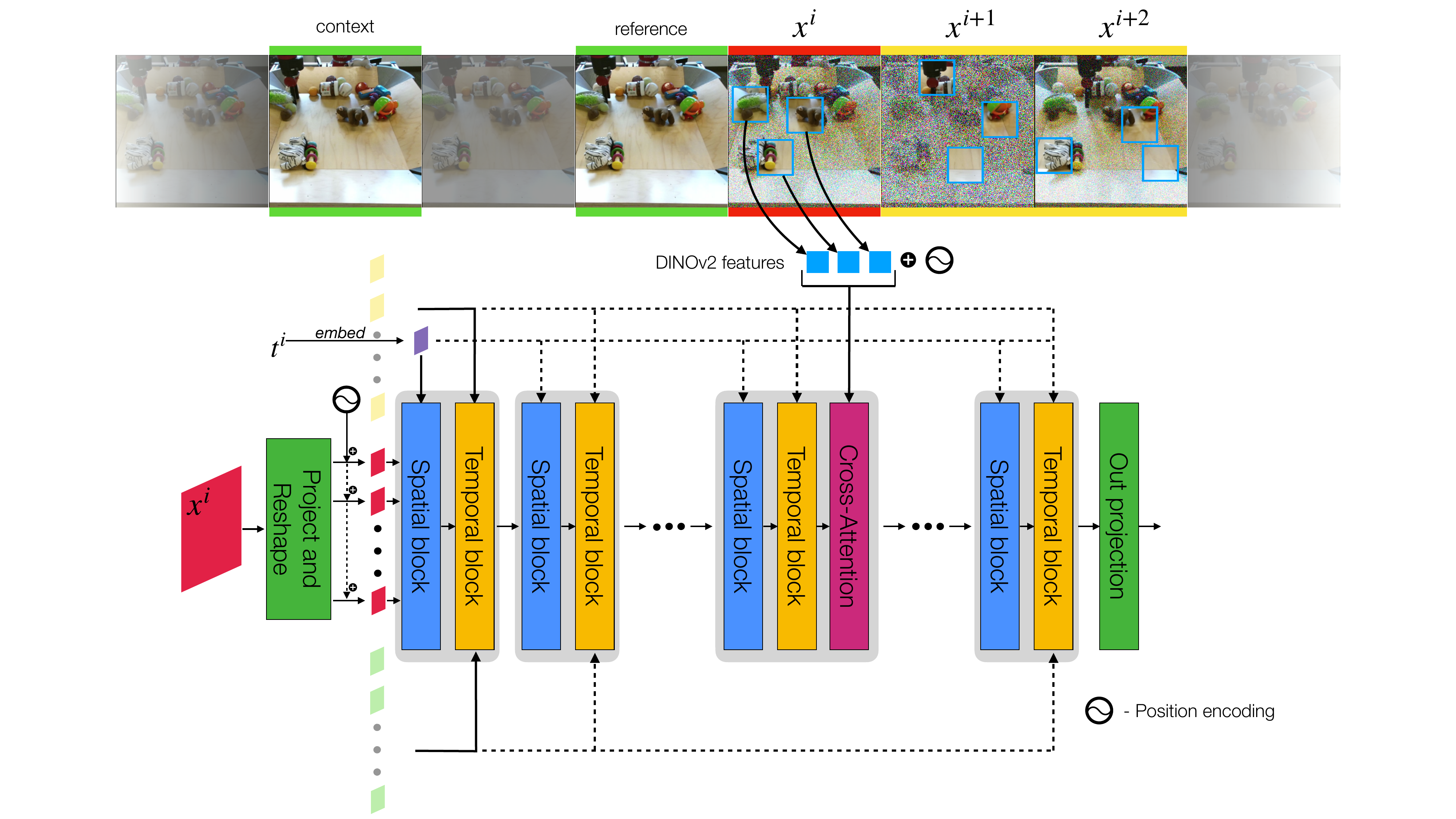}
    \caption{Overall pipeline of \methodName. The model takes all the colored frames and processes them equally and in parallel. The pipeline for a single frame ($x^i$, in red) is illustrated. \methodName is trained to predict the denoising direction for the future frames ($x^{i:i+2}$) in the CFM~\cite{lipman2022flow} framework conditioned on the past frames (\textit{context} and \textit{reference}) and sparse random sets of DINOv2~\cite{Oquab2023DINOv2LR} features. The frames communicate with each other via the Temporal Blocks while being separately processed by the Spatial Blocks. The controls are incorporated through Cross-Attention.}
    \label{fig:pipeline}
\end{figure*}

The goal of controllable video generation is to learn the following conditional distribution
\begin{align}\label{eq:dist}
    p(x^{n + 1 : n + k}\;|\; x^{1:n}, a^{n + 1:n + k}), 
\end{align}
where $x^i \in \mathbb{R}^{3 \times H \times W}, \; i = 1, \dots, n + k$ is a sequence of RGB video frames, $a^i$ is the control at time $i$, and $x^{i:j}$ ($a^{i:j}$) denotes the set of consecutive video frames (controls) between times $i$ and $j\ge i$. Following \cite{Davtyan_2023_ICCV} we model the distribution in eq.~\ref{eq:dist} as a denoising process in the conditional flow matching (CFM) formulation of diffusion \cite{lipman2022flow}. In CFM, the model $v_t(x^{n+1:n+k}_t\;|\; x^{1:n}, a^{n+1:n+k}, \theta)$
with parameters $\theta$ is trained to approximate the direction of the straight line that connects independently sampled noise and data points:
\begin{align}\label{eq:obj}
    \theta^* = \arg\min_\theta {\mathbb{E}} \,\left\|v_t - x^{n+1:n+k}_1 + (1 - \sigma_m) x^{n+1:n+k}_0 \right\|^2_2.
\end{align}
Here the expectation is with respect to $t, x_0, x_1$, and $a$, where $t$ is a random timestamp sampled from $U[0, 1]$, $x^i_0 \sim {\cal N}(0, I)$, $x^{1:n+k}$ and $a^{n+1:n+k}$ are sampled from the dataset, and $x^i_t = t x^i + (1 - (1 - \sigma_m) t) x^i_0$ with a small $\sigma_m \approx 10^{-7}$. It can be shown that integrating from $t=0$ to $t=1$ the following ODE
\begin{equation}\label{eq:ode}
\begin{split}
    \dot X_t &= v_t(X_t\;|\; x^{1:n}, a^{n+1:n+k}, \theta^*), \\
    p(X_0) &= {\cal N}(0, I),
\end{split}
\end{equation}
where $X_t$ denotes the window of noisy future frames $x^{n+1:n+k}_t$, leads to $p(X_1) \approx p(x^{n + 1 : n + k}\;|\; x^{1:n}, a^{n + 1:n + k})$. That is, to sample from the probability density function in eq.~\eqref{eq:dist} one can sample Gaussian noise at time 0 and gradually denoise it by following the trajectories of eq.~\eqref{eq:ode} till time $t=1$. For more details, please refer to \cite{lipman2022flow}.

As suggested in \cite{Davtyan_2023_ICCV} we relax the computational complexity of conditioning on the past frames by distributing the conditioning over the flow integration steps. This allows to effectively condition each step only on two past observations: the previous one $x^n$ (the reference) and one uniformly sampled from the past $x^c$ (the context). Thus, we work with the model $v_t(x^{n+1:n+k}_t\;|\; x^n, x^c, a^{n+1:n+k}, \theta), \; c \sim U\{1, \dots, n - 1\}$. At inference, $c$ is also randomized at each integration step of the ODE. This allows the model to observe all the past frames, while keeping the computational costs constant.

We also observed that decoupling the noise levels of separate frames during training is beneficial to the quality of the generated frames. \emph{I.e.}, we set $t = (t_1, \dots, t_k)$ with independently sampled $t_i$ that are not necessarily equal to each other and $X_t = x_t^{n+1:n+k} = (x_{t_1}^{n + 1}, \dots, x_{t_k}^{n + k})$. This allows the noisy frames to exploit shared information (\eg, the background) in the less noisy frames and leads to improved training. Besides this, the model with decoupled $t$ is more flexible at inference, where the denoising can be done within a sliding window of growing noise levels. However, here we leave exploring this capability to future work and instead focus on the model's controllability. A similar technique was proposed for motion synthesis~\cite{zhang2023tedi} and for video diffusion~\cite{ruhe2024rolling}.

Lastly, before we proceed to the definition of our control format, we would like to point out an important requirement for our model. To enable the  synthesis of a scene given only the controls (also referred to as ``\emph{from scratch}''), we drop the conditioning on the context frame for 50\% of the training. Moreover, we drop the reference frame 20\% of the time during which we drop the context frame. By simultaneously training our model under different conditioning settings, we endow it with the ability to both predict videos from some initial frames and to generate them from scratch.

\subsection{Controlling Scene Composition over Time via Sparse Features}

\begin{figure*}[t]
    \centering
    \includegraphics[width=0.65\linewidth, trim=5cm 14cm 5cm 14cm, clip]{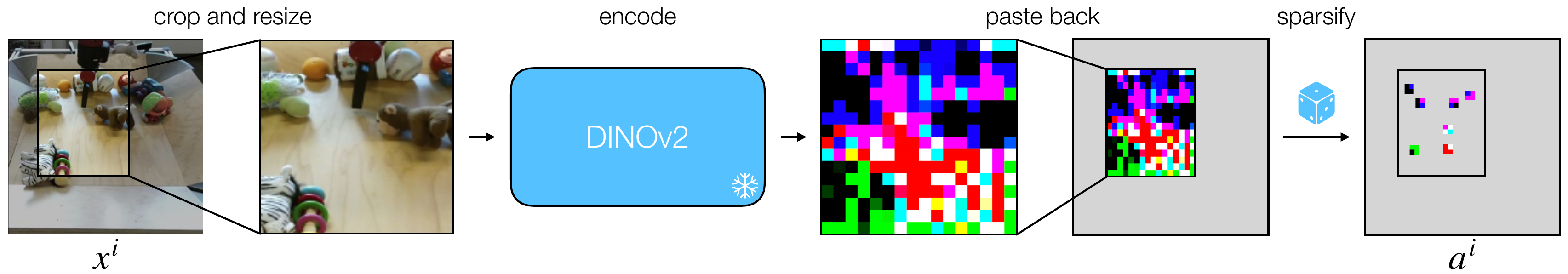}
    \caption{The process of selecting controls for conditioning. The image is first cropped and resized to $224\times224$ resolution that can be fed to DINOv2 to obtain the spatial tokens. Those are then pasted back to the original location of the crop and sparsified. This is done to prevent overfitting to the position information that is present in DINOv2 features. Besides this, calculating the features on the crops of the image makes the model scale invariant. That is, at inference we are able to copy objects from the background and paste them to the foreground and vice versa. The model should be able to automatically figure out how to scale the objects according to their target position in the scene as well as how to add other position-related textures (e.g. shadows).}
    \label{fig:controls}
\end{figure*}

Our goal is to build a video generation model in which the controls can serve both as motion guidance for the video prior and as scene generation specifications. The latter control means that when we omit the conditioning on the past, the model has to generate and animate the scene from scratch. The control input describes the scene in terms of objects (or parts thereof) and their positions in space and time. In contrast to prior work that is either limited to only scene generation~\cite{Epstein2022BlobGANSD,Hudson2021CompositionalTF} or uses supervision~\cite{Huang2023FinegrainedCV,Ma2023TrailBlazerTC}, instead of leveraging composite separate solutions for scene generation and animation~\cite{Bruce2024GenieGI, Wang2024BoximatorGR}, we seek for a unified control that can simultaneously solve both problems in an unsupervised way. 

To this end, we propose to use a sparse set of DINOv2~\cite{Oquab2023DINOv2LR} spatial tokens from the last $l$ ViT~\cite{vit} layers as the controls $a^i$. More precisely, for each future frame $x^i$, each spatial token from the $16\times 16$ grid $f^i \in \mathbb{R}^{d\times 16 \times 16}$ of its DINOv2 encodings is assigned a probability $\pi\in[0,1]$ to be selected for control. Then, a random $16 \times 16$ mask $m^i \in \mathbb{R}^{16 \times 16}$ is sampled from the corresponding Bernoulli distribution. This mask consists of 0s and 1s, where 1s stand for the selected tokens. The control is then calculated as
\begin{align}
    a^i = m^i \cdot f^i + (1 - m^i) \cdot \text{[MSK]},
\end{align}
where [MSK] $\in \mathbb{R}^{d \times 1 \times 1}$ is a  trainable token. 

By conditioning on sparse DINOv2 features and training for video prediction (generation) we gain two major advantages. 1) The controls are unified and data-agnostic. This means that, during testing, a scene can be constructed by positioning DINOv2 features extracted from unseen images. These images can even belong to domains not present in the training set, as we later demonstrate. Object motion can be specified by simply adjusting the position of the same features in future frames. 2) The controls are abstract enough to enable the prediction of  changes in the appearance and location of moving objects across multiple frames by using the same features. In contrast, if RGB patches at the pixel level were used as controls, the model could overfit by relying on the texture details of specific object instances to recover their exact positions. Indeed, in the ablations we show that the controllability suffers from too much information in the control tokens. This motivates us to use features rather than raw color patches. By default, unless otherwise stated, we leverage the ViT-S/14 version of DINOv2.

Note that our model is general enough to leverage other feature extractors that satisfy the properties mentioned above, such as CLIP~\cite{radford2021learning}. However, we empirically find DINOv2 to perform better. Moreover, we favor the use of DINOv2, in contrast to CLIP, because it was trained in an unsupervised manner, which makes CAGE fully unsupervised. We ablate the choice of DINOv2 as the feature extractor in the supplementary material.

\subsection{Scale and Position Invariance}

Even though DINOv2 features are quite abstract, they preserve information about the original position of the corresponding  patches in the input images~\cite{Yang2023EmerNeRFES}. Naively conditioning on those features would lead to overfitting to such positional information and hence achieve limited controllability over the locations of the objects (see Sec.~\ref{sec:ablations} and the supplementary material). In order to mitigate this issue, we propose to feed the features in such a way that the identities of the objects would be preserved, but their positional information would be destroyed. To do so, we calculate the features on random crops instead of whole images. This approach ensures that the position of objects within the crop varies throughout training, encouraging the model to disregard positional information. This procedure is illustrated in Fig.~\ref{fig:controls}.

\subsection{Out of Distribution Controls}

YODA~\cite{davtyan2024learn} demonstrated the ability to generalize to out of distribution (o.o.d.) control. In fact, YODA can move background objects in the BAIR dataset~\cite{Ebert2017SelfSupervisedVP}. In the training set, background objects only move when pushed by the robotic arm. This is accomplished by balancing two objectives during training: 1) learning the video prior, which defines how objects should move when no controls are specified, and 2) learning to accurately follow the provided controls. The right balance is the outcome of tuning many components, among which the most important ones, according to~\citet{davtyan2024learn}, are the number of the controls and where they are sampled. In contrast, we propose to keep the training simple and  rely on the compositionality. Our method has no restrictions on where the controls can be sampled from. Instead, we propose to find the balance at test time by utilizing classifier-free guidance~\cite{ho2022classifier} to allow generalization to the o.o.d. controls. More precisely, we modify the estimated vector field with
\begin{align}
    v_t = (1 + w) \cdot &v_t(X_t\;|\; x^{1:n}, a^{n+1:n+k}, \theta^*) \\
    - w \cdot &v_t(X_t\;|\; x^{1:n}, \varnothing, \theta^*),
\end{align}
where $w \geq 0$ is the guidance strength. The larger $w$ the more the model relies on the control (to move background objects) rather than on the video prior (not to move background objects).
Fig.~\ref{fig:ood_bair} shows some examples of o.o.d. controls in BAIR.

\subsection{Training Details and Architecture}
\label{sec:arch}

For computational efficiency, as proposed in~\cite{rombach2022high}, \methodName works in the latent space of a pre-trained VQGAN~\cite{esser2021taming}. All frames $x^{n:n+k}$ as well as $x^c$ are separately encoded to latents. Each latent is a feature map of shape $c\times h\times w$. These latents are first reshaped to $(h\cdot w)\times c$. Then a random set of tokens (but the same across time) is dropped from each of the latent codes, leaving $m = \lfloor (1 - r) \cdot h \cdot w \rfloor$ tokens per latent. Here $r$ is the masking ratio that is typically equal to $0.4$ in our experiments. The remaining tokens are then concatenated in the first dimension to form a single sequence of $(k+2) \cdot m$  tokens. This sequence is then passed to $v_t$, which we model as a Transformer~\cite{Vaswani2017AttentionIA} that consists of 9 blocks of alternating spatial and temporal attention layers. Spatial layers are allowed to only attend to the tokens within a given frame, while temporal layers observe all the tokens. As suggested in~\cite{Hatamizadeh2023DiffiTDV}, $t$ is embedded in the architecture as a time-depended bias that is added to the queries, keys and values in all attention layers. We add learnable spatial position encodings to all the tokens before feeding them to $v_t$. The same position encodings are later added to the control tokens $a^{n+1:n+k}$. For the video indices temporal layers employ relative position encodings~\cite{relative_posenc}. 5 middle blocks of the Transformer additionally incorporate cross-attention layers, where tokens from the $i$-th frame attend to $a^i$.  
The overall pipeline of \methodName is depicted in Fig.~\ref{fig:pipeline}.


\section{Experiments}
\label{sec:experiments}

In this section we conduct a series of experiments on several datasets to highlight different capabilities of \methodName and demonstrate its superiority over the prior work.

\subsection{Data}
\label{sec:data}

We test our model on 3 datasets. Ablations are conducted on the \textbf{CLEVRER}~\cite{clevrer} dataset. This dataset consists of 10k training videos capturing multiple synthetic objects colliding into each other and interacting on a flat surface. This dataset combines simplicity of the objects with quite interesting interactions and supports compositionality making it a great field for experimentation. We further test \methodName on the \textbf{BAIR}~\cite{Ebert2017SelfSupervisedVP} dataset, which is a dataset containing 44k clips of a real robotic arm manipulating diverse set of objects on a table. This dataset provides more complex objects and interactions and is suitable for demonstrating generalization to o.o.d. controls. Finally, we test our model on real egocentric videos from the \textbf{EPIC-KITCHENS}~\cite{epic} dataset. For making the training more resource efficient, we restrict ourselves to a single person/kitchen from the dataset, namely P03.

\subsection{Ablations}\label{sec:ablations}

We start by ablating different components of \methodName on the CLEVRER~\cite{clevrer} dataset. For ablations we train \methodName with a smaller batch size of 16 samples. However, later we show that the performance of the model scales accordingly with larger batch sizes.

In order to select the best configuration of the model, we assess its controllability in terms of scene composition. To this end, we have annotated 128 images from the test set of the CLEVRER dataset by selecting 4-6 random patches of the same object in an image. Then, the model is fed as input a real frame and the selected patches shifted to a random target location, and outputs a generated frame. The quality of the generated frame is measured with a combination of two metrics: FID~\cite{fid} for image realism and CTRL, which we propose for measuring control accuracy (\ie, how much the generated frame follows the input control). CTRL is the ratio of $S$ over $D$. $S$ is high when the object in the real image has moved to the desired target destination. It is measured as the cosine distance between normalized DINOv2 features at the initial position in the real frame and those at the target location in the generated frame. $D$ instead is high when the object does not move. It is measured as the cosine distance between normalized DINOv2 features at the initial position in the real frame and those at the same initial location but in the generated frame. Finally, CTRL is high when the model is highly controllable and low otherwise. Empirically we observe that both $S$ and $D$ are lower bounded by some positive number, which makes CTRL a valid metric. We test different configurations of the model and demonstrate that the variant with 1 DINOv2 layer, $\pi = 0.1$, $k = 3$, with randomized $t_i$ and scale/position invariance performs the best on CLEVRER (see Table~\ref{tab:ablations}). We provide visual examples to highlight the limited control in the ablated models in the supplementary material.

\begin{table}[t]
    \centering
\begin{tblr}{
  rows = {belowsep=0pt},
  column{1-4} = {wd=0.2cm, halign=c, valign=m, colsep=8pt},
  column{4} = {wd=0.7cm, halign=c, valign=m, colsep=6pt},
  column{5} = {wd=0.8cm, halign=c, valign=m, colsep=12pt},
  column{6-11} = {halign=c},
  vline{6} = {0.7pt},
  hline{1,10} = {1pt,solid},
}
     $l$ & $\pi$ & $k$ &
     {\small random $t_i$} &
     {\small scale/pos inv.} &  CTRL$\uparrow$ & FID$\downarrow$ \\
     \hline
     1 & 0.1 & 3 & \checkmark &  & 1.306 & 12.88 \\
     1 & 0.1 & 1 & \checkmark & \checkmark & 1.529 & 9.85 \\
     1 & 0.1 & 3 & & \checkmark & \underline{1.596} & 5.82 \\
     1 & 0.9 & 3 & \checkmark & \checkmark & 1.464 & 6.19 \\
     1 & 0.5 & 3 & \checkmark & \checkmark & 1.566 & 5.19 \\ 
     1 & 0.01 & 3 & \checkmark & \checkmark & 0.782 & 6.11 \\
     4 & 0.1 & 3 & \checkmark & \checkmark & 1.084 &  5.24 \\
     2 & 0.1 & 3 & \checkmark & \checkmark & 1.500 & \textbf{4.95} \\
     \SetRow{gray9}
     1 & 0.1 & 3 & \checkmark & \checkmark & \textbf{1.615} & \underline{4.98} \\ 
\end{tblr}
\caption{Ablations of model components with respect to compositional scene generation quality. The controllability score (CTRL) and the image quality (FID) are evaluated. The row corresponding to the full model is highlighted in gray. For the CTRL metric the average of 5 runs is reported.}\label{tab:ablations}
\end{table}

\subsection{Quantitative Results}
\label{sec:quantitative_results}

\noindent\textbf{CLEVRER.} Employing the optimal settings identified in the ablation studies, we train \methodName with the full batch size of 64 samples and report the results in Tab.~\ref{tab:clevrer}. Besides the scene composition metrics studied in the ablations, following~\citet{davtyan2024learn}, we reconstruct 15 frames of a video from a single initial one conditioned on the control for the first generated frame. We report LPIPS~\cite{lpips}, PSNR, SSIM~\cite{ssim} and FVD~\cite{fvd}. Table~\ref{tab:clevrer} shows that \methodName generates videos of better quality compared to prior work.

\begin{table}[t]
    \centering
    \footnotesize
    \newcommand{\curSpace}{1mm}
    \begin{tabular}{@{}l@{\hspace{\curSpace}}c@{\hspace{\curSpace}}c@{\hspace{\curSpace}}c@{\hspace{\curSpace}}c@{\hspace{\curSpace}}c@{\hspace{\curSpace}}c@{}}
        \toprule 
        Method & LPIPS$\downarrow$ & PSNR$\uparrow$ & SSIM$\uparrow$ & FID$\downarrow$ & FVD$\downarrow$ & CTRL$\uparrow$\\
        \hline
            YODA & \textbf{0.126} & \textbf{30.07} & 0.93 & 5.7 & 70 & -\\
            \methodName \textit{(ours)} & 0.166 & 30.02 & \textbf{0.98} & \textbf{4.3} & \textbf{62} & 1.667 \\
        \bottomrule
    \end{tabular}
    \caption{Comparison with YODA~\cite{davtyan2024learn} on the \emph{CLEVRER} dataset. 15 frames generated from a single one. Control is provided for the first generated frame.}\label{tab:clevrer}
\end{table}

\noindent\textbf{BAIR.} Following prior work by Menapace \textit{et al.}~\cite{menapace2021playable}, we trained \methodName on the BAIR dataset~\cite{Ebert2017SelfSupervisedVP}. We autoregressively generated 29 frames starting from a given initial frame and employing various controls. For feature extraction on BAIR we found that it is optimal to use the 2 last layers of ViT-B/14 variant of DINOv2. The experiment was conducted on the  test videos from BAIR, utilizing an Euler solver with 50 steps as our ODE solver to ensure precise temporal evolution. We assessed the quality of the generated frames by reporting LPIPS~\cite{lpips} scores (between the generated images and the ground truth frames) and FID~\cite{fid} scores. More importantly, we utilized FVD~\cite{fvd} to quantify the quality of the generated motions and the consistency of the frames. To verify the generalization capability of our conditioning scheme, we evaluated the same model under three distinct settings (results shown in Tab.~\ref{tab:bair}). In the 10\% control setting, we encoded future frames using the same DINOv2 model employed during training and conditioned the model on 10\% of these features, randomly sampled with $\pi = 0.1$, which mirrors the training setting. To assess the robustness of our model, we also evaluated it with only 1\% of the future features (\ie by setting $\pi = 0.01$, resulting in 1-2 features per frame on average), achieving superior FID and FVD scores compared to other models. Lastly, unlike the previous two settings, which were non-causal (i.e., conditioned on the features of the ground truth future frames), we experimented with a setting where features from the first frame were propagated to subsequent frames using the flow determined by assessing the cosine similarity between future patch features and the first frame's features. We applied a high cosine similarity threshold of 0.95 to ensure the accuracy of the flows and maintain control percentage comparability with the training setting. This setting approximates the use of the model at inference, and the experiment shows that \methodName performs better than the prior work even under this domain gap between the training and the test times.


\begin{table}[t]
\centering
\footnotesize
\begin{tabular}{l@{\vspace{0.5mm}}crr}
    \toprule
        Method & LPIPS$\downarrow$ & FID$\downarrow$ & FVD$\downarrow$\\
    \hline
        CADDY~\cite{menapace2021playable} & 0.202 & 35.9 & 423 \\
        Huang et al.~\cite{huang2022layered} \\
        \makecell[l]{\qquad\emph{positional}}  & 0.202 & 28.5 & 333 \\
        \makecell[l]{\qquad\emph{affine}} & 0.201 & 30.1 & 292 \\
        \makecell[l]{\qquad\emph{non-param}} & 0.176 & 29.3 & 293 \\
        GLASS~\cite{davtyan2022glass} & 0.118 & 18.7 & 411 \\
        YODA~\cite{davtyan2024learn} \\
        \makecell[l]{\qquad\textit{5 controls}} & \underline{0.112} & 18.2 & 264 \\
        \makecell[l]{\qquad\textit{1 control}} & 0.142 & 19.2 & 339 \\
    \hline
        \methodName \textit{(ours)} \\
        \makecell[l]{\qquad\textit{10\% controls}} & \textbf{0.107} & \textbf{6.4} & \textbf{136} \\
        \makecell[l]{\qquad\textit{1\% controls}} & 0.149 & \underline{7.2} & \underline{169} \\
        \makecell[l]{\qquad\textit{tracking}} & 0.194 & 9.1 & 214 \\
    \bottomrule
\end{tabular}
\caption{Evaluation on the \emph{BAIR} dataset.} \label{tab:bair}
\end{table}

\begin{table}[t]
    \centering
    \footnotesize
    \begin{tabular}{@{}l@{\hspace{1mm}}c@{\hspace{1.9mm}}c@{\hspace{1.9mm}}c@{\hspace{1.9mm}}c@{}}
    \toprule
         Method & \makecell{GT} & \makecell{from GT \\dist.} & \makecell{o.o.d.\\ S} & \makecell{o.o.d.\\ L} \\
        \hline
        CADDY~\cite{menapace2021playable} & 6.29 & - & - & - \\
        GLASS~\cite{davtyan2022glass} & 1.75 & 25.00 & 25.07 & 30.06 \\
        YODA~\cite{davtyan2024learn}& 1.41 & 1.77 & 2.01 & 3.37 \\

        \hline
        \methodName \textit{(ours)} & 2.74 & 2.93 & 3.57 & 5.45 \\
        \methodName \textit{(ours)} w/ pos. interpolation & 2.29 & 2.33 & 2.85 & 4.77 \\ 
    \bottomrule
    \end{tabular}
    \caption{Optical flow error in pixels of the control applied to the robotic arm in the \emph{BAIR} dataset. Average of 5 runs.}\label{tab:control_baselines}
\end{table}

\begin{table}[t]
    \centering
    \footnotesize
    \begin{tabular}{@{}l@{\vspace{0.5mm}}c@{\hspace{1.9mm}}c@{\hspace{1.9mm}}c@{\hspace{1.9mm}}c@{}}
        \toprule 
        Method & LPIPS$\downarrow$ & PSNR$\uparrow$ 
        & FID$\downarrow$ & FVD$\downarrow$\\
        \hline
            YODA~\cite{davtyan2024learn} & 0.436 & 16.22 &
            152 & 664 \\
            \methodName \textit{(ours)} & \textbf{0.283} & \textbf{22.25} & 
            \textbf{95} & \textbf{393} \\
        \bottomrule
    \end{tabular}
    \caption{Evaluation of the generated videos on the \emph{EPIC-KITCHENS} dataset.}\label{tab:epic}
\end{table}

Even though methods with different controls are technically not comparable, other than for video quality, and different controllability metrics incorporate biases towards favoring certain controls, for completeness, we measure the controllability score proposed in~\cite{davtyan2024learn} and report it in Tab.~\ref{tab:control_baselines}. This score measures the discrepancy in pixels between the intended control (which in \cite{davtyan2024learn} is an optical flow vector that is applied to an object) and the optical flow estimated with a pre-trained optical flow network (RAFT~\cite{teed2020raft} in this case) between the initial and the generated frames. As done in the ablations, we annotate 128 frames from the BAIR dataset with the locations of the robot patches and apply random shifts to the patches to assess the controllability. Following~\cite{davtyan2024learn}, we report the controllability score in 4 different settings: 1) the shift is estimated from the ground truth future frame 2) the shift is generated with a uniform direction and the norm sampled from the ground truth distribution, which is ${\cal N}(7.19, 5.12)$ 3) o.o.d. S - the same as 2), but the norm comes from ${\cal N}(10, 0.1)$ 4) o.o.d. L - the same as 2), but the norm comes from ${\cal N}(20, 0.1)$. \methodName performs slightly worse than~\cite{davtyan2024learn} in this experiment. However, it is important to note that the metric used here is measured in pixels, while our method operates with patches arranged on a fixed grid. To overcome the limitation of grid-aligned patches and capture shifts smaller than the grid step, we propose interpolating the position encodings added to features before using them as controls. This technique is implemented in a zero-shot use of our pre-trained model and demonstrates better controllability compared to the vanilla implementation. Finally, the highest score of \methodName, which is 4.77, is less than half the patch size, which is $16\times 16$. Based on that, we can claim the controllability of our method.

\noindent\textbf{EPIC-KITCHENS.} We perform the same reconstruction experiment as before for the EPIC-KITCHENS dataset to show that our model works in even more realistic settings. We use the model configuration from BAIR and autoregressively generated 15 frames from 1 conditioned on controls estimated from the future ground truth frames. We also trained YODA~\cite{davtyan2024learn} on the same data using the official codebase~\footnote{\url{https://github.com/araachie/yoda}} and evaluated it the same way. The results are in Tab.~\ref{tab:epic}.





\begin{figure}[t]
    \centering
    \newcommand\curWidth{1.6cm}
    \begin{tabular}{@{}c@{\hspace{0.5mm}}c@{\hspace{0.5mm}}c@{}}
        $w = 0.0$ & $w = 0.0$ & $w = 7.0$ \\
        \animategraphics[width=\curWidth]{7}{Figures/bair_ood_ball_robot/image_0000}{0}{9} & \animategraphics[width=\curWidth]{7}{Figures/bair_ood_ball_ng/image_0000}{0}{9} & \animategraphics[width=\curWidth]{7}{Figures/bair_ood_ball/image_0000}{0}{9} \\
        \animategraphics[width=\curWidth]{7}{Figures/bair_ood_zebra_robot/image_0000}{0}{9} & 
        \animategraphics[width=\curWidth]{7}{Figures/bair_ood_zebra_ng/image_0000}{0}{9} & \animategraphics[width=\curWidth]{7}{Figures/bair_ood_zebra/image_0000}{0}{9} \\

    \end{tabular}
    \caption{The effect of CFG on the generalization to out of distribution controls on the BAIR dataset. While the robotic arm can be controlled with no guidance ($w = 0.0$), with larger $w$ the model is also able to move the background objects not moving on their own in the training data. Click on the images  to play them as videos in Acrobat Reader.}
    \label{fig:ood_bair}
\end{figure}

\begin{figure}[t]
    \centering
    \newcommand\curWidth{1.4cm}
    \begin{tabular}{@{}c@{\hspace{0.5mm}}c@{\hspace{0.5mm}}c@{\hspace{0.5mm}}c@{\hspace{0.5mm}}c@{\hspace{0.5mm}}c@{}}
        source image & control & \multicolumn{3}{c}{generated sequence$\rightarrow$} \\
         \includegraphics[height=\curWidth, trim=1.0cm 0cm 1.0cm 0cm, clip]{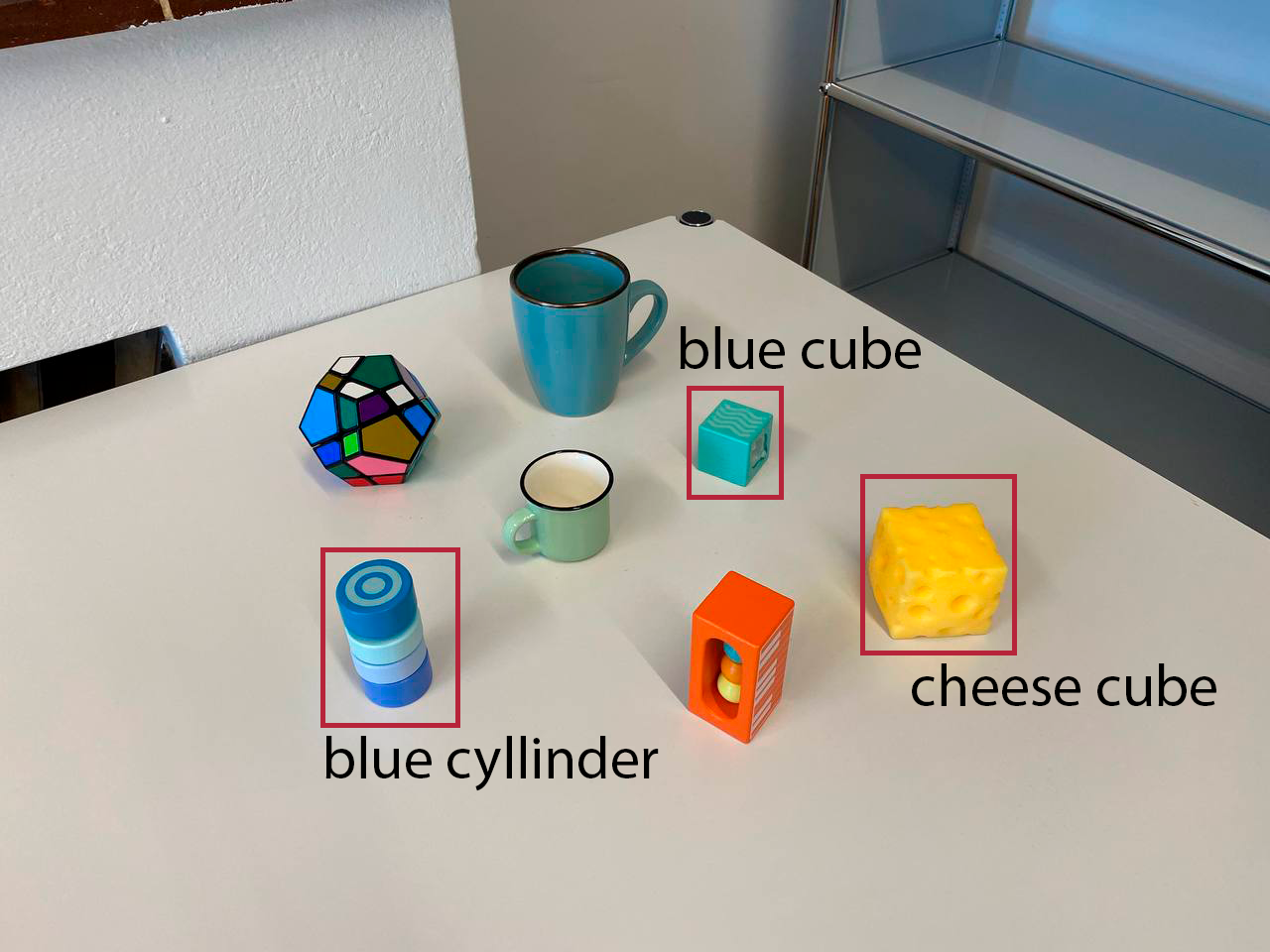} & \includegraphics[width=\curWidth]{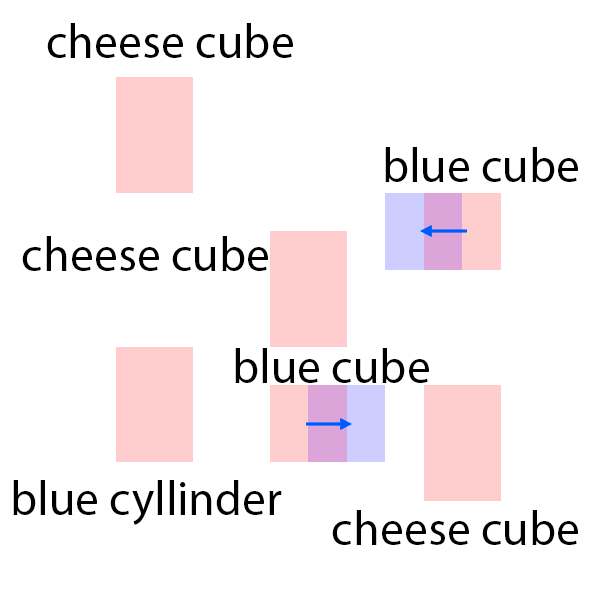} & 
         \animategraphics[width=\curWidth]{7}{Figures/cl_cross_2/image_0000}{0}{8} & 
         \includegraphics[width=\curWidth]{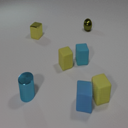} & 
         \includegraphics[width=\curWidth]{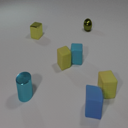} \\
         \includegraphics[height=\curWidth, trim=1.0cm 0cm 1.0cm 0cm, clip]{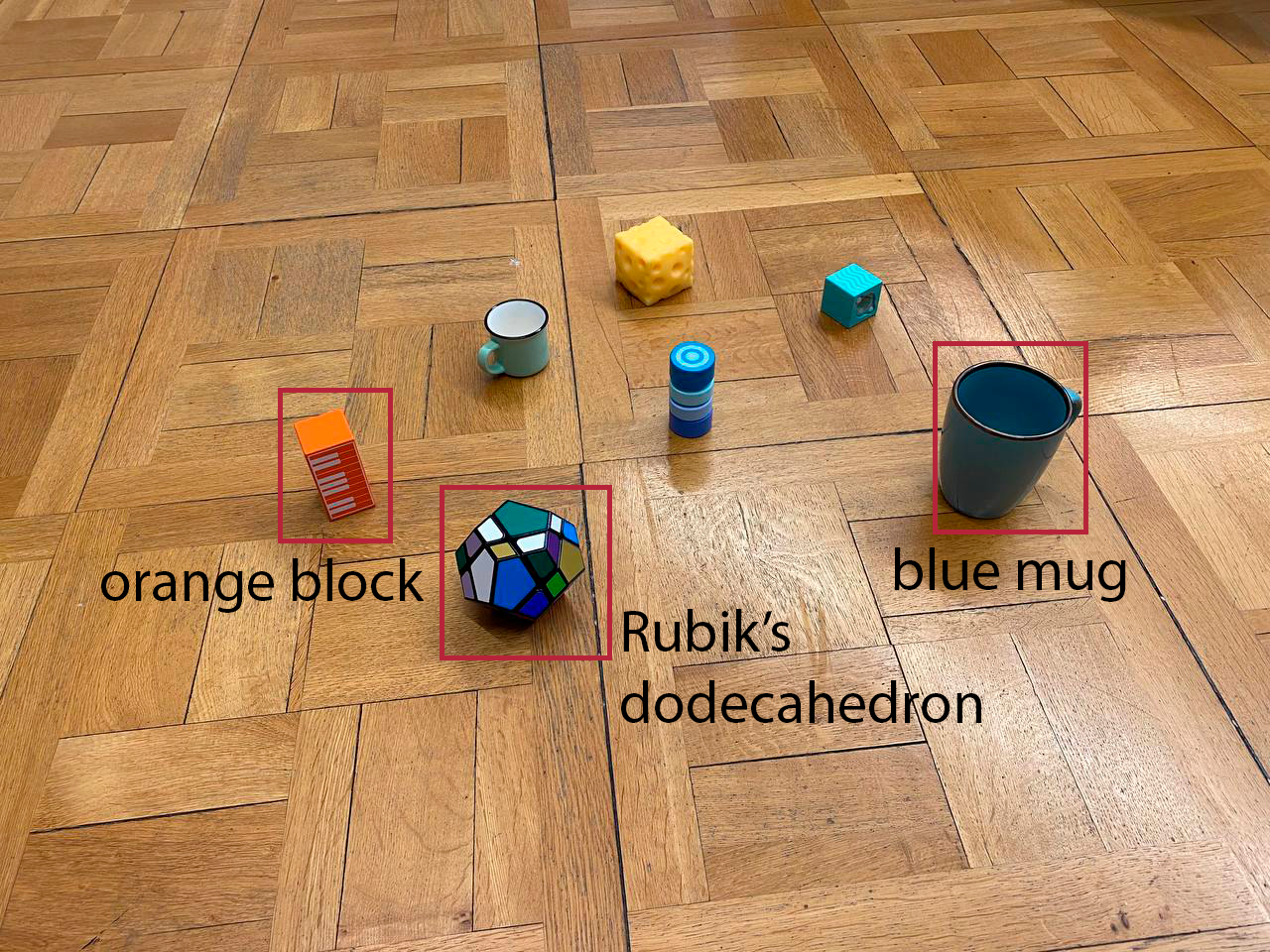} & \includegraphics[width=\curWidth]{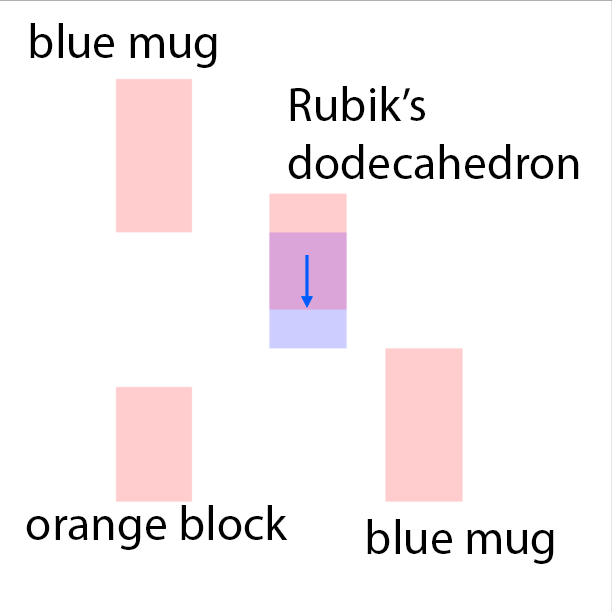} & 
         \animategraphics[width=\curWidth]{7}{Figures/cl_cross_1/image_0000}{0}{8} & 
         \includegraphics[width=\curWidth]{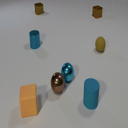} & 
         \includegraphics[width=\curWidth]{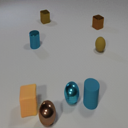} \\ 
    \end{tabular}
    \caption{Examples of cross-domain transfer. The features of the objects from images in the first column are borrowed to compose and animate the scenes in the CLEVRER dataset. 
    Notice how \methodName resolves the domain gap and performs a reasonable transfer of the objects (in terms of shapes and colors). However some objects with irregular shape and texture, such as the Rubik's dodecahedron, may turn to multiple objects when transfered. Click on the first images in the generated sequences to play them as videos in Acrobat Reader.}
    \label{fig:cross}
\end{figure}

\subsection{Qualitative Results}
\label{sec:qualitative_results}

In this section, we show some qualitative results to visually illustrate various capabilities of \methodName (more, including videos, can be found in the supplementary material as well as on the project's website). 

The main contribution of our model is the ability to both compose and animate scenes using a unified control. This is demonstrated in Fig.~\ref{fig:comp}. With the trained model one is able to select object patches from different images and compose them into a novel scene. This is achieved by extracting the DINOv2 features from the selected patches and placing them in the indicated locations to form the control that is fed to the model. Additionally, one may also specify the motion of the objects by moving the patches in the future frames. \methodName carefully adjusts the appearance of the objects to the locations of the features in the control map, while preserving the objects' identities. Fig.~\ref{fig:ood_bair} depicts some out of distribution controls on BAIR. As previously discussed, using larger $w$ in classifier-free guidance allows us to adapt to o.o.d. control, such as moving background objects in BAIR.
Finally, in Fig.~\ref{fig:cross} we demonstrate the ability of the model to generalize to cross-domain scenarios. We select the features for control from images from a different domain and transfer them to the data domain the model was trained on (e.g. real objects to CLEVRER objects). This is possible because of our specific choice of conditioning the generation on DINOv2 features that were trained on a large open image dataset and hence are abstract and data-agnostic.


\section{Conclusion and Limitations}
\label{sec:conclusion}


We introduced \methodName, an unsupervised method for controllable video generation capable of composing and animating scenes using objects from other frames or out-of-domain images. This is achieved by conditioning the video generation on sparse DINOv2 spatial tokens describing object appearance and spatiotemporal location. We demonstrated our method's capabilities through various experiments.
Due to limited computing resources, our experiments were conducted on relatively small datasets, precluding direct comparison with state-of-the-art large-scale video generation models trained on extensive annotated data.
However, we believe that the training pipeline, the architecture, and the unsupervised nature of the considered problem enable the scalability of our model and hope that the ideas explored in our work will pave the path towards large-scale unsupervised foundation models for controllable video generation.

\section{Acknowledgements}

This work was supported by grant 188690 of the Swiss National Science Foundation.

\appendix

\section{Appendix}

\setcounter{secnumdepth}{1}

\section{Additional Details}\label{seq:add_details}

\noindent\textbf{Training.} Following~\cite{rombach2022high}, we train \methodName in the latent space of pretrained VQGAN~\cite{esser2021taming} models. For the BAIR~\cite{Ebert2017SelfSupervisedVP} and CLEVRER~\cite{clevrer} datasets we used the pretrained autoencoders from the official repository of YODA~\cite{davtyan2024learn}~\footnote{\url{https://github.com/araachie/yoda}}. For the EPIC-KITCHENS~\cite{epic} dataset, we trained the autoencoder using the official implementation of VQGAN~\footnote{\url{https://github.com/CompVis/taming-transformers}}. The training specifications of VQGAN can be found in Table~\ref{tab:vqgan}. The model for the EPIC-KITCHENS dataset was trained on a single kitchen/person, namely \texttt{P03}. We used 49 videos for training and left out 1 video (\texttt{P03\textunderscore 28}) for the testing purposes.
All models except for the ablations are trained for 100k iterations with batch size 64 distributed across 4 Nvidia RTX 3090 GPUs. The models for the ablations are trained on a single GPU with batch size 16. All models are trained in mixed-precision (fp16) using AdamW~\cite{Loshchilov2017DecoupledWD} optimizer with the square root learning rate schedule with the largest learning rate equal to 1e-4 and 5k warmup steps. Weight decay of 5e-6 is used.

\noindent\textbf{Inference.} Once the model is trained, we generate videos in an autoregressive manner, subsequently denoising windows of future frames conditioned on previously generated ones. In order to generate the current window, one needs to start by sampling a noise tensor from the standard normal distribution and then integrate the flow ODE from 0 to 1 using a numerical ODE solver (we used the \texttt{euler} solver with 20-40 steps).

The code, the pretrained models, and the demo that provides a convenient access to our models can be found in the project's github repository \url{https://github.com/araachie/cage}.

\begin{table*}[t]
    \centering
    \begin{tabular}{@{}r@{\hspace{0.5mm}}|@{\hspace{5mm}}c@{}}
          & EPIC-KITCHENS $256\times 256$~\cite{epic} \\
    \hline
        embed\textunderscore dim & 8 \\
        n\textunderscore embed  & 16384 \\
        double\textunderscore z  & False \\ 
        z\textunderscore channels & 8 \\
        resolution & 256 \\
        in\textunderscore channels & 3 \\
        out\textunderscore ch  & 3 \\
        ch  & 128 \\
        ch\textunderscore mult  & [1,1,2,2,4] \\
        num\textunderscore res\textunderscore blocks  & 2\\
        attn\textunderscore resolutions & [16] \\
        dropout & 0.0 \\
        disc\textunderscore conditional  & False \\
        disc\textunderscore in\textunderscore channels & 3 \\
        disc\textunderscore start & 20k \\
        disc\textunderscore weight & 0.8 \\
        codebook\textunderscore weight & 1.0 \\
    \end{tabular}
    \caption{Training configuration of VQGAN~\cite{esser2021taming} for the EPIC-KITCHENS~\cite{epic} dataset.}
    \label{tab:vqgan}
\end{table*}

\section{Additional Qualitative Results}\label{seq:add_results}
\begin{figure*}[t]
    \footnotesize
    \centering
    \newcommand{\curWidth}{0.13\linewidth}
    \begin{tabular}{ccccc}
         source & control & MAE & CLIP & DINOv2 \\
         \includegraphics[width=\curWidth]{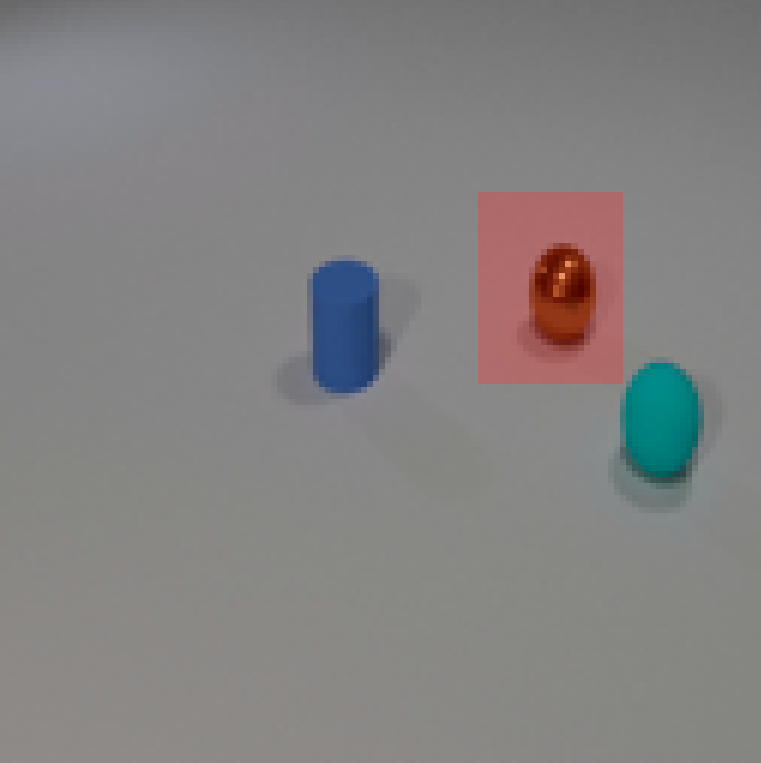} & 
         \includegraphics[width=\curWidth]{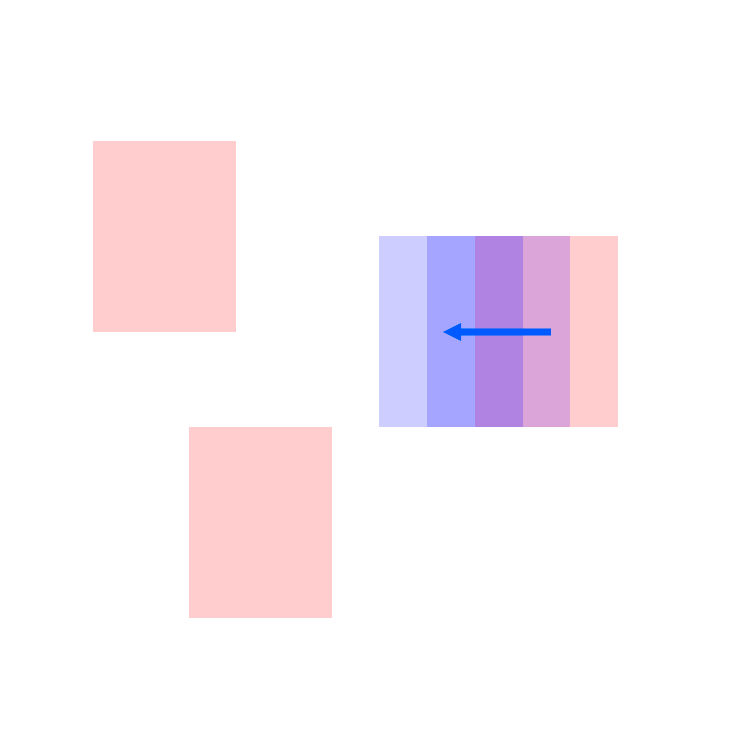} &  
         \animategraphics[width=\curWidth]{7}{Figures/clevrer_mae_1/image_0000}{0}{8} &
         \animategraphics[width=\curWidth]{7}{Figures/clevrer_clip_1/image_0000}{0}{8} &
         \animategraphics[width=\curWidth]{7}{Figures/clevrer_dinov2_1/image_0000}{0}{8} \\
          & 
         \includegraphics[width=\curWidth]{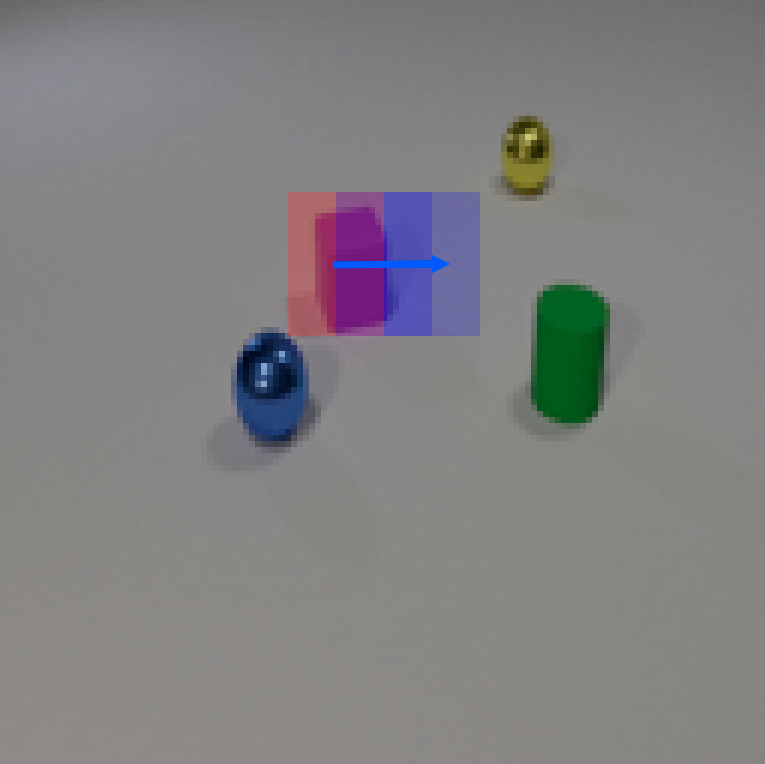} &  
         \animategraphics[width=\curWidth]{7}{Figures/clevrer_mae_2/image_0000}{0}{8} &
         \animategraphics[width=\curWidth]{7}{Figures/clevrer_clip_2/image_0000}{0}{8} &
         \animategraphics[width=\curWidth]{7}{Figures/clevrer_dinov2_2/image_0000}{0}{8} \\
    \end{tabular}
    \caption{Ablation of the feature extractor used to train \methodName. Top row: composition and animation. Bottom row: animation from a given frame. Click on the images in columns 3-5 to play them as videos in Acrobat Reader.}
    \label{fig:feats_ablation}
\end{figure*}

\begin{figure*}[t]
    \centering
    \newcommand\curWidth{0.14\linewidth}
    \begin{tabular}{@{}c@{\vspace{0.5mm}}c@{\vspace{0.5mm}}c@{\vspace{0.5mm}}c@{\vspace{0.5mm}}c@{\vspace{0.5mm}}c@{}}
        \multicolumn{6}{c}{generated sequence $\rightarrow$} \\ 
        \animategraphics[width=\curWidth]{7}{Figures/epic_up/image_0000}{0}{9} & \includegraphics[width=\curWidth]{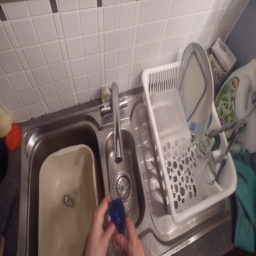} & \includegraphics[width=\curWidth]{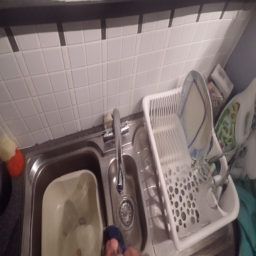} & \includegraphics[width=\curWidth]{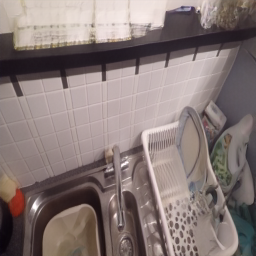} & \includegraphics[width=\curWidth]{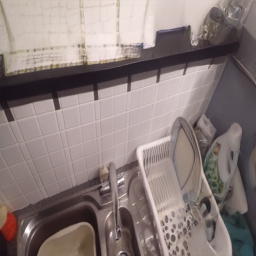} & \includegraphics[width=\curWidth]{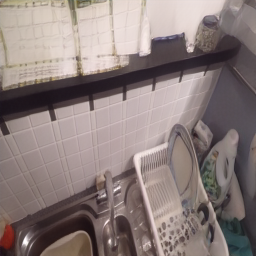} \\
        \animategraphics[width=\curWidth]{7}{Figures/epic_right/image_0000}{0}{9} & \includegraphics[width=\curWidth]{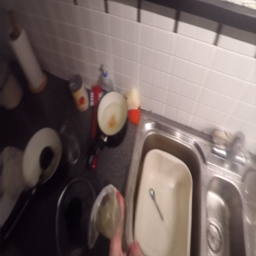} & \includegraphics[width=\curWidth]{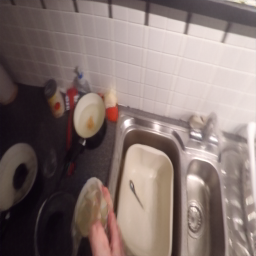} & \includegraphics[width=\curWidth]{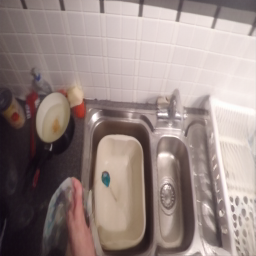} & \includegraphics[width=\curWidth]{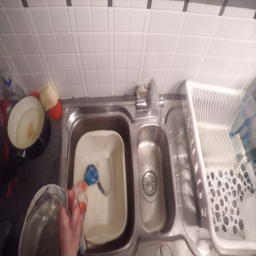} & \includegraphics[width=\curWidth]{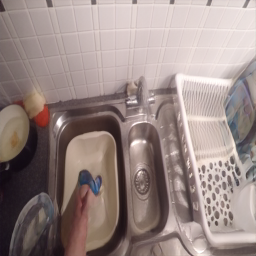} \\
        \animategraphics[width=\curWidth]{7}{Figures/epic_left/image_0000}{0}{9} & \includegraphics[width=\curWidth]{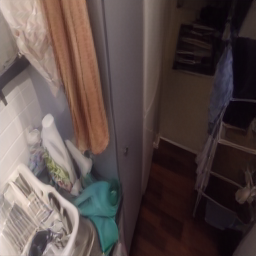} & \includegraphics[width=\curWidth]{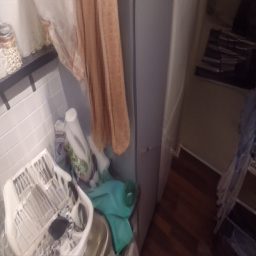} & \includegraphics[width=\curWidth]{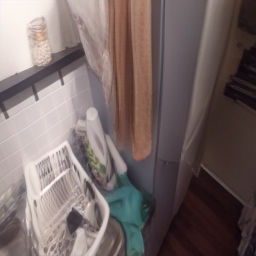} & \includegraphics[width=\curWidth]{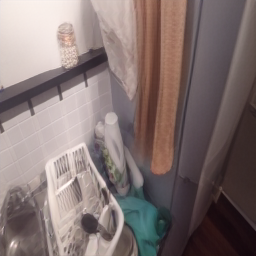} & \includegraphics[width=\curWidth]{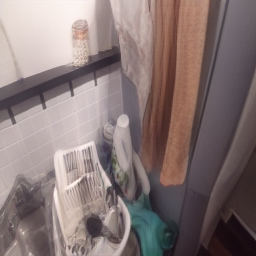} \\
        \animategraphics[width=\curWidth]{7}{Figures/epic_forward/image_0000}{0}{9} & \includegraphics[width=\curWidth]{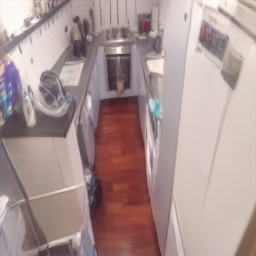} & \includegraphics[width=\curWidth]{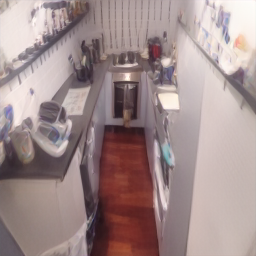} & \includegraphics[width=\curWidth]{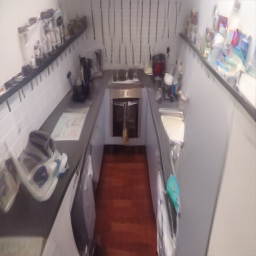} & \includegraphics[width=\curWidth]{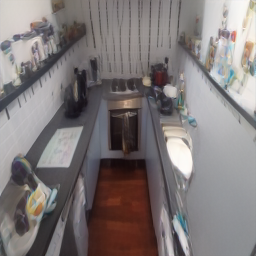} & \includegraphics[width=\curWidth]{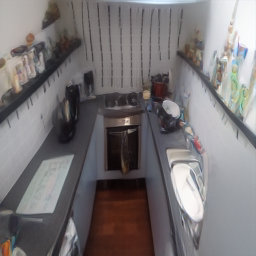} \\
        \animategraphics[width=\curWidth]{7}{Figures/epic_hand/image_0000}{0}{9} & \includegraphics[width=\curWidth]{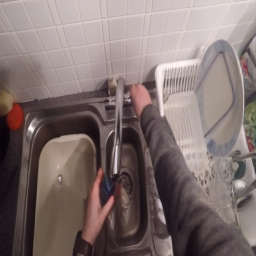} & \includegraphics[width=\curWidth]{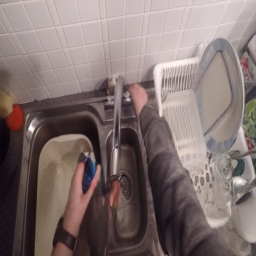} & \includegraphics[width=\curWidth]{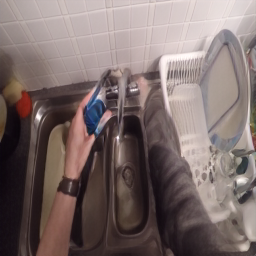} & \includegraphics[width=\curWidth]{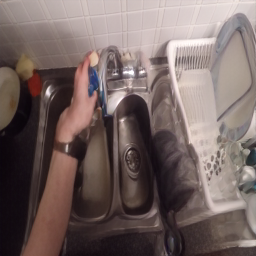} & \includegraphics[width=\curWidth]{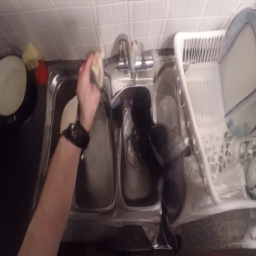} \\
    \end{tabular}
    \caption{Generated videos on the EPIC-KITCHENS dataset. \methodName is able to model such motions as ego-motion or the movement of the hands. The controls are superimposed with the images in the first column. Click on the images in the first column to play them as videos in Acrobat Reader. Notice that the blue arrows are shown only to illustrate the intended motion and are not part of the control signal, which consists only of features from the selected patches that are moved in the direction pointed with the blue arrows.}
    \label{fig:epic}
\end{figure*}

In this section we provide more videos generated with \methodName. All the results in the main paper and this supplementary material are obtained using images from the test sets of the corresponding datasets, \ie unseen during training. For convenience, we designed a webpage that presents the videos in a structured way. To view it, please open  \url{https://araachie.github.io/cage/}. The figures in this pdf on the other hand show the quality of separate frames that cannot be assessed with the gifs from the webpage. The videos in Fig.~\ref{fig:feats_ablation} and Fig.~\ref{fig:epic} are only embedded in this PDF and are not available on the website. The other figures contain only images, and the corresponding videos, which duplicate the results from those figures, are included on the website.

\begin{figure*}[t]
    \centering
    \newcommand\curWidth{0.13\linewidth}
    \newcommand\curSpace{1mm}
    \begin{tabular}{@{}c@{\hspace{\curSpace}}c@{\hspace{\curSpace}}c@{\hspace{\curSpace}}c@{\hspace{\curSpace}}c@{}}
         \makecell{selected \\object} & \makecell{target\\ layout} & \makecell{full\\ model} & \makecell{w/o \\ cropping} & $l = 4$ \\
         \includegraphics[width=\curWidth]{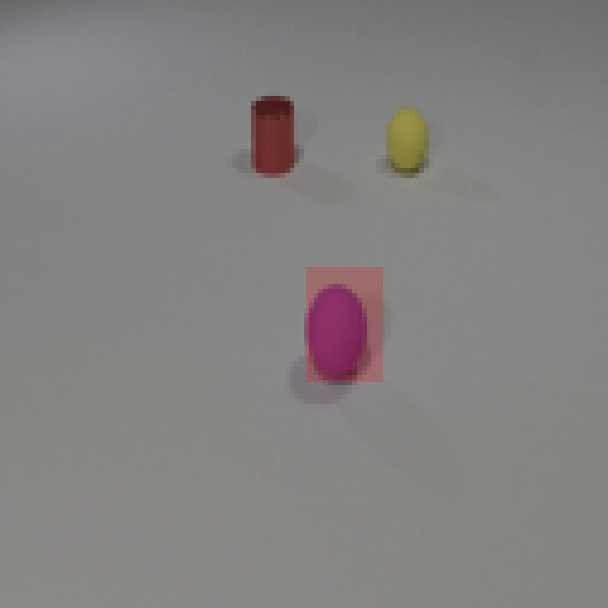} & \includegraphics[width=\curWidth]{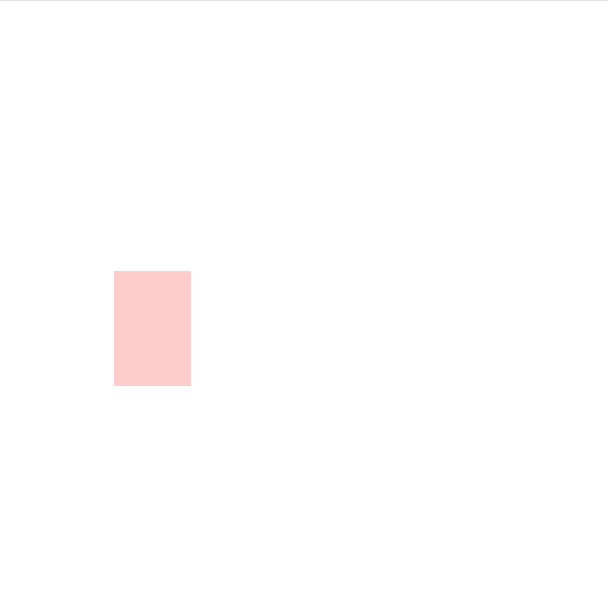} & \includegraphics[width=\curWidth]{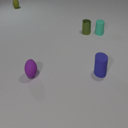} & \includegraphics[width=\curWidth]{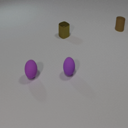} & \includegraphics[width=\curWidth]{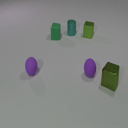} \\
         \includegraphics[width=\curWidth]{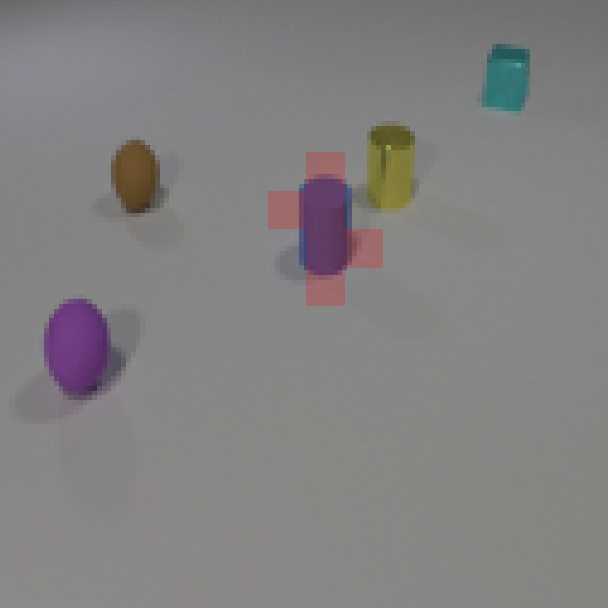} & \includegraphics[width=\curWidth]{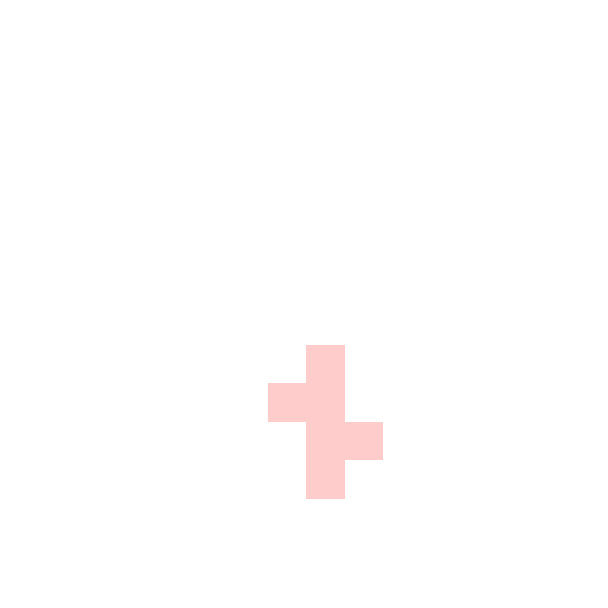} & \includegraphics[width=\curWidth]{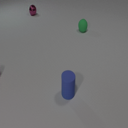} & \includegraphics[width=\curWidth]{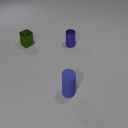} & \includegraphics[width=\curWidth]{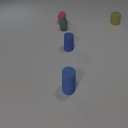} \\
    \end{tabular}
    \caption{Illustration of the effect of omitting calculation of the features on random crops (4th column) and using more DINOv2 layers (5th column). Those models tend to overfit to either position information or redundant texture information in the features, which results in limited controllability. In constrast to those models, the full model does not keep the moved object in the original location.}
    \label{fig:ablations}
\end{figure*}

\noindent\textbf{Ablations.} Fig.~\ref{fig:ablations} visually supports the ablations in the main paper and shows the effect of keeping more information in the control features. As discussed in the main paper, we opted for DINOv2 features because they satisfy the following criteria: 1) they were trained in an unsupervised way, 2) they provide some invariance (to small pertubations like color, illumination, scale, pose); thus, they can represent the same moving object across time, 3) they can generalize across domains and represent semantically similar objects. We have also tested CAGE with features from MAE~\cite{he2022masked} and CLIP~\cite{radford2021learning} on CLEVRER. MAE doesn't satisfy the second criterion, as it was trained with the reconstruction loss and preserves too much information about the input. It produces poor results with CTRL=1.001. CAGE trained with CLIP features succeeds in generating controllable videos with CTRL=1.302, but is worse than DINOv2 (CTRL=1.667) and does not satisfy the first criterion to be used in our fully unsupervised pipeline. See the qualitative results in Fig.~\ref{fig:feats_ablation}.

\begin{figure*}[t]
\centering
\newcommand\curWidth{0.12\linewidth}
\begin{tabular}{@{}c@{\hspace{0.5mm}}c@{\hspace{0.5mm}}c@{\hspace{0.5mm}}c!{\vrule width 1pt}c@{\hspace{0.5mm}}c@{\hspace{0.5mm}}c@{}}
    1st source & 2nd source & 3rd source & control & \multicolumn{3}{c}{generated sequence $\rightarrow$} \\
     \includegraphics[width=\curWidth]{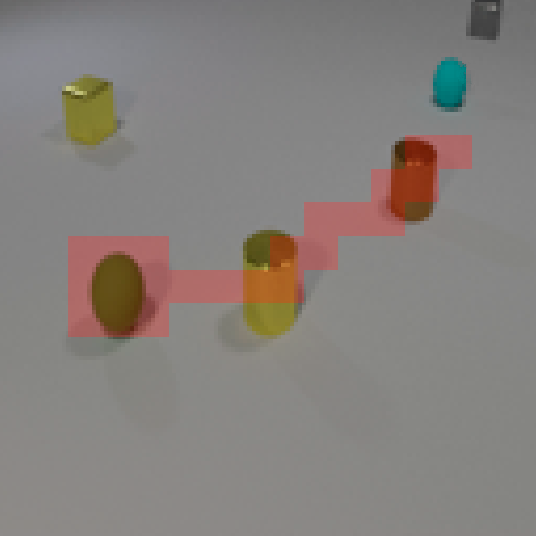} &
     \includegraphics[width=\curWidth]{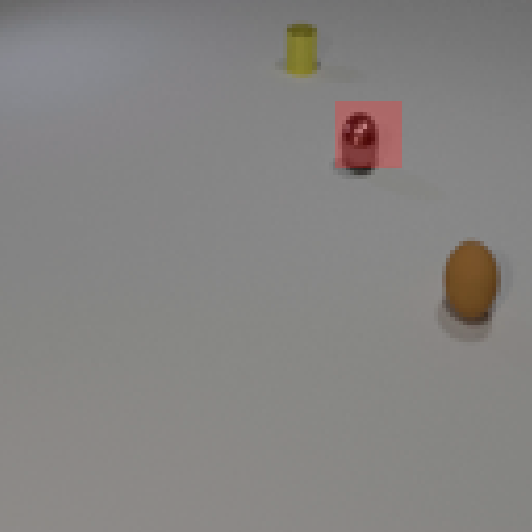} & 
     \includegraphics[width=\curWidth]{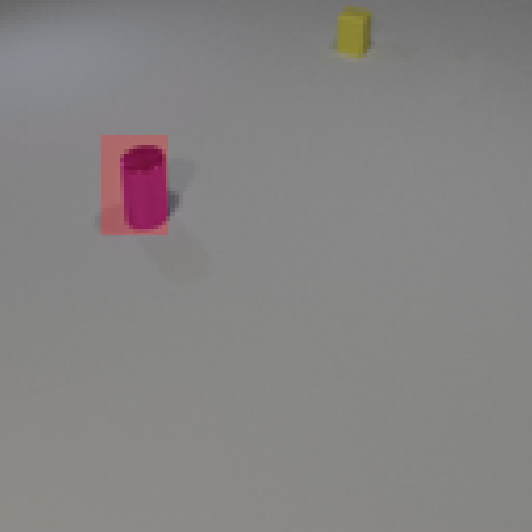} &
     \includegraphics[width=\curWidth]{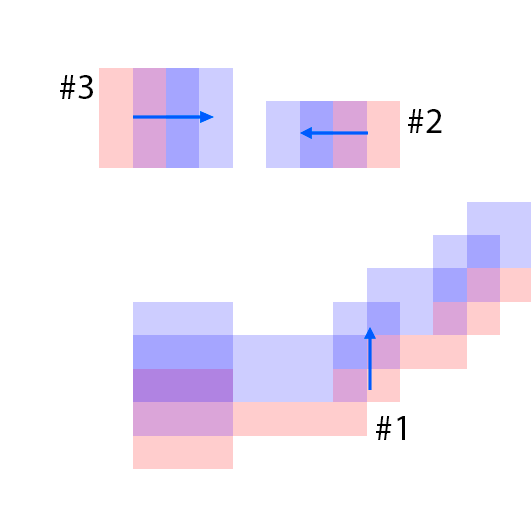} & 
     
     \includegraphics[width=\curWidth]{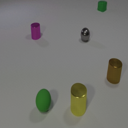} &
     \includegraphics[width=\curWidth]{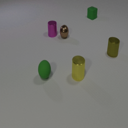} & 
     \includegraphics[width=\curWidth]{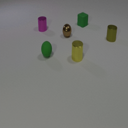} \\
     \includegraphics[width=\curWidth]{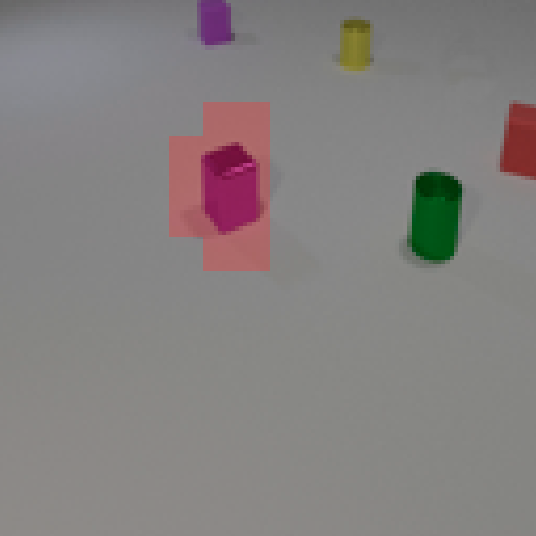} &
     \includegraphics[width=\curWidth]{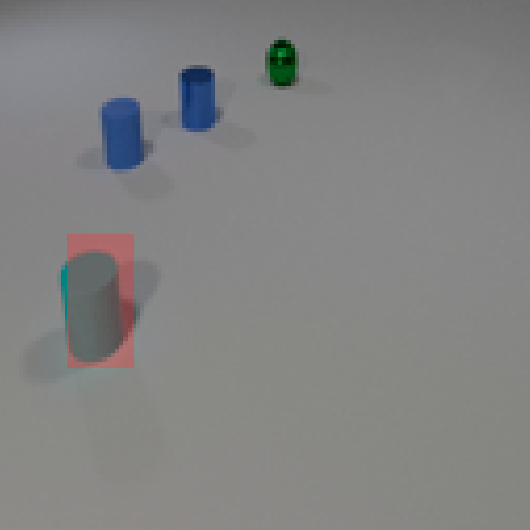} & 
     \includegraphics[width=\curWidth]{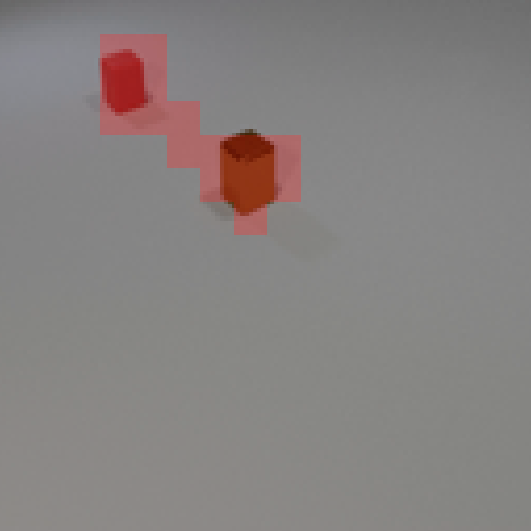} &
     \includegraphics[width=\curWidth]{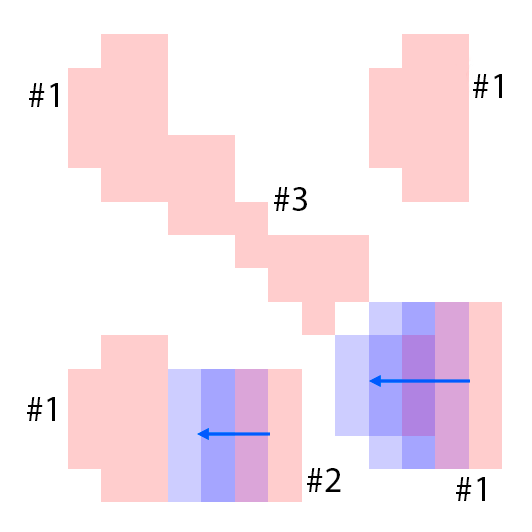} & 
     
     \includegraphics[width=\curWidth]{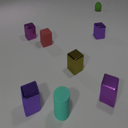} & 
     \includegraphics[width=\curWidth]{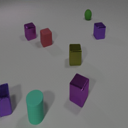} & 
     \includegraphics[width=\curWidth]{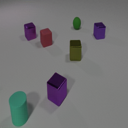} \\
     \includegraphics[width=\curWidth]{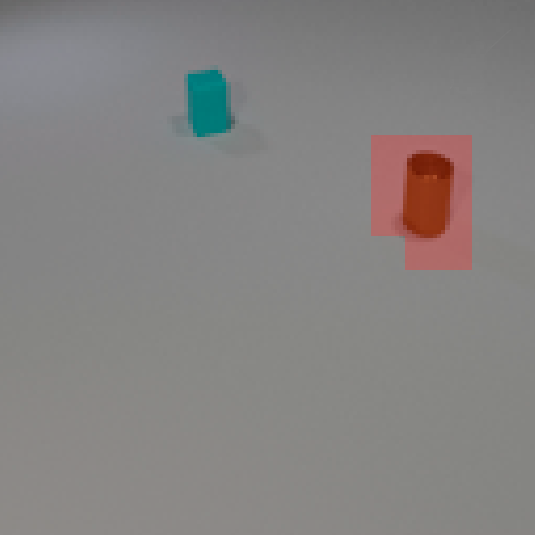} &
     \includegraphics[width=\curWidth]{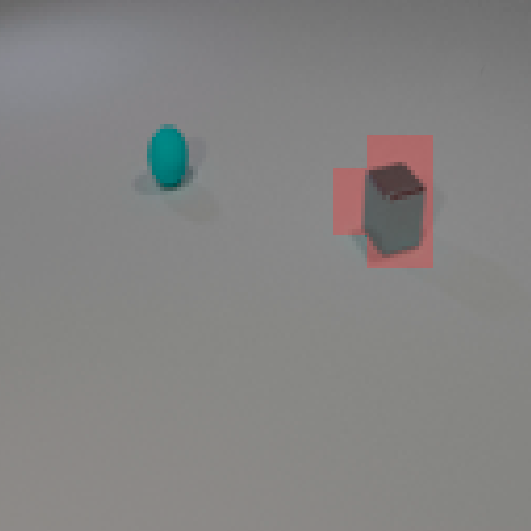} & 
     \includegraphics[width=\curWidth]{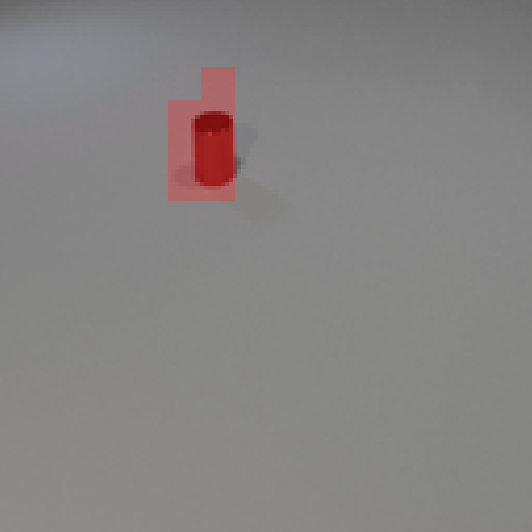} &
     \includegraphics[width=\curWidth]{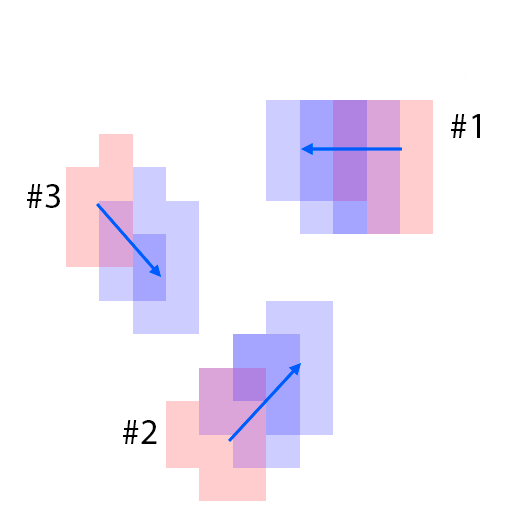} & 
     
     \includegraphics[width=\curWidth]{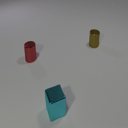} & 
     \includegraphics[width=\curWidth]{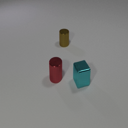} & 
     \includegraphics[width=\curWidth]{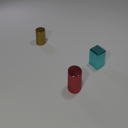} \\
     \includegraphics[width=\curWidth]{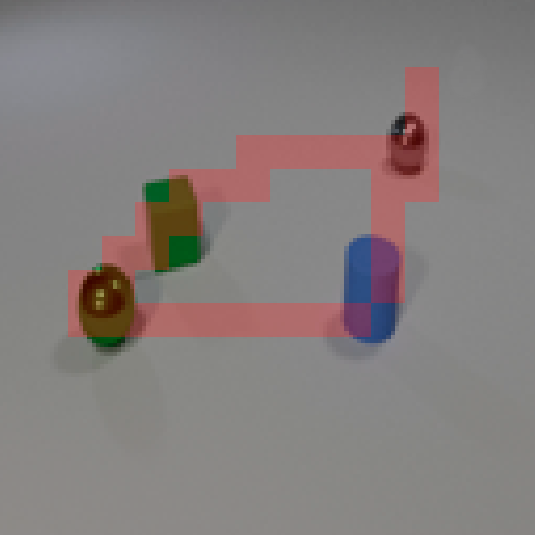} &
     \includegraphics[width=\curWidth]{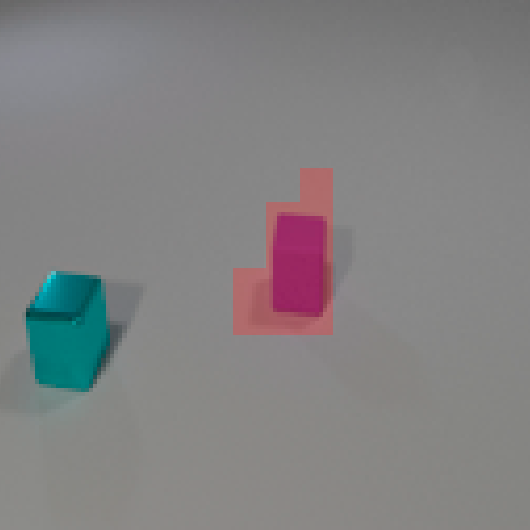} & &
     \includegraphics[width=\curWidth]{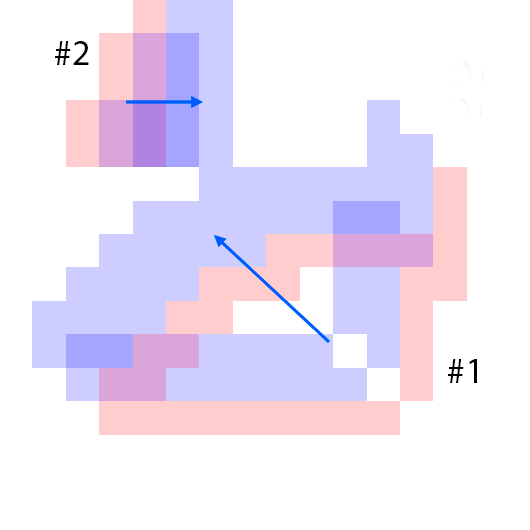} & 
     
     \includegraphics[width=\curWidth]{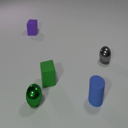} & 
     \includegraphics[width=\curWidth]{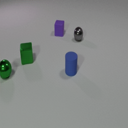} & 
     \includegraphics[width=\curWidth]{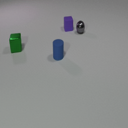} \\
     \includegraphics[width=\curWidth]{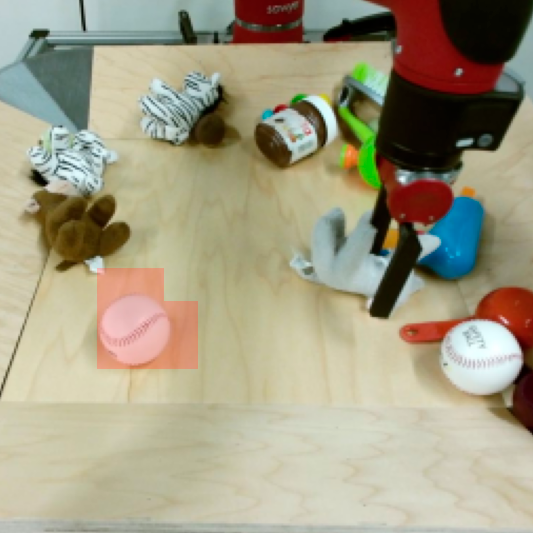} &
     \includegraphics[width=\curWidth]{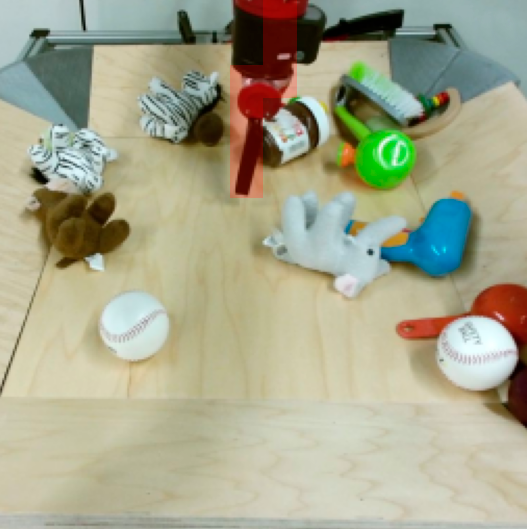} & &
     \includegraphics[width=\curWidth]{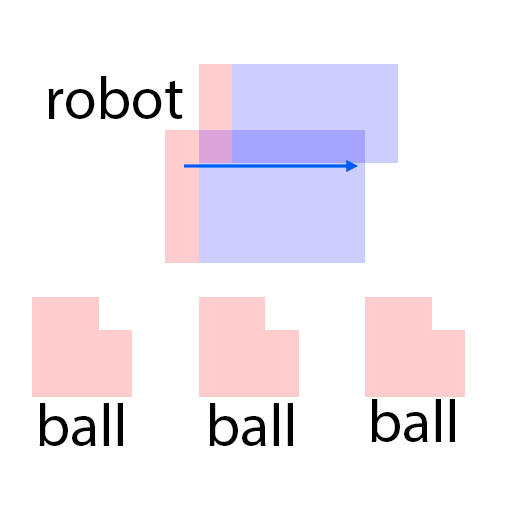} & 
     
     \includegraphics[width=\curWidth]{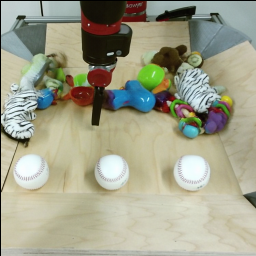} & 
     \includegraphics[width=\curWidth]{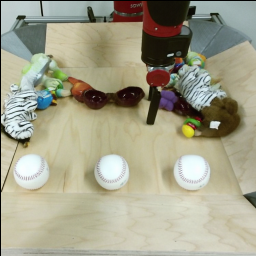} & 
     \includegraphics[width=\curWidth]{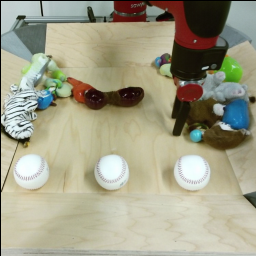} \\
     \includegraphics[width=\curWidth]{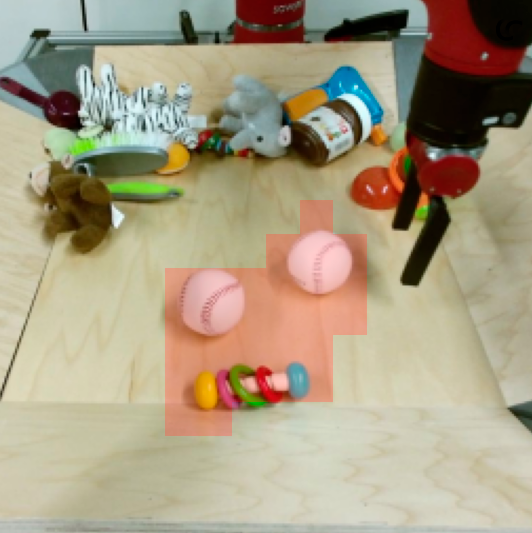} &
     \includegraphics[width=\curWidth]{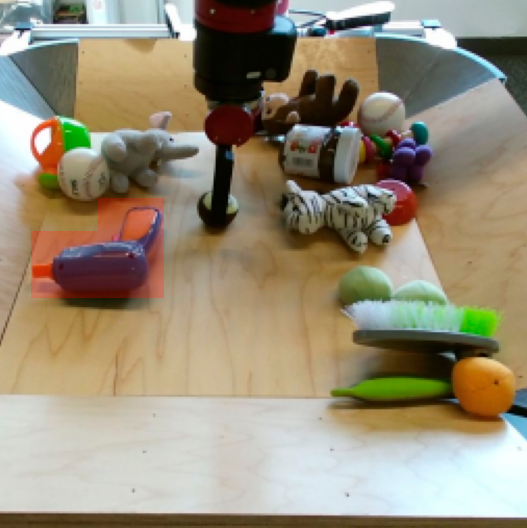} & &
     \includegraphics[width=\curWidth]{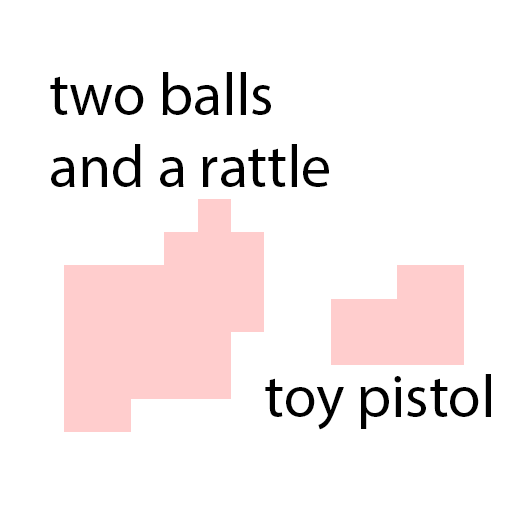} & 
     
     \includegraphics[width=\curWidth]{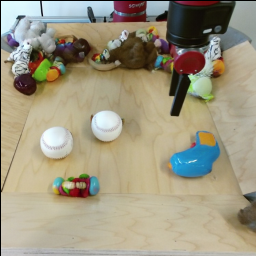} & 
     \includegraphics[width=\curWidth]{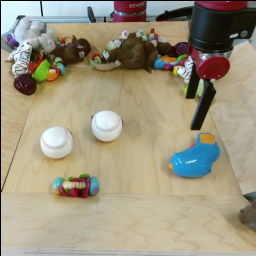} & 
     \includegraphics[width=\curWidth]{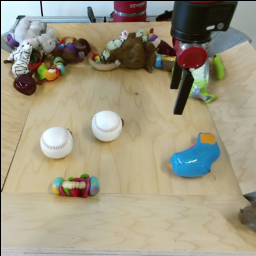} \\
     \includegraphics[width=\curWidth]{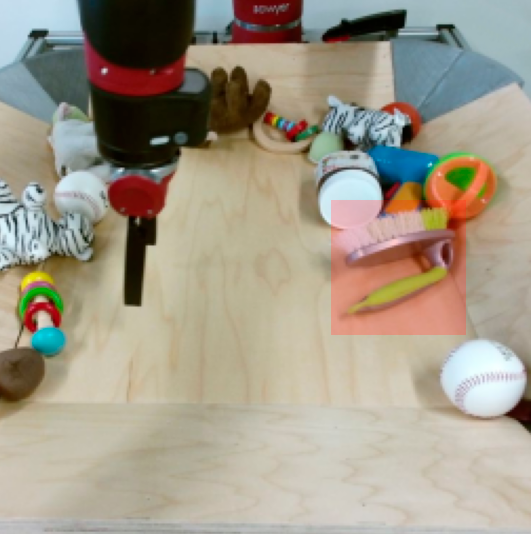} &
     \includegraphics[width=\curWidth]{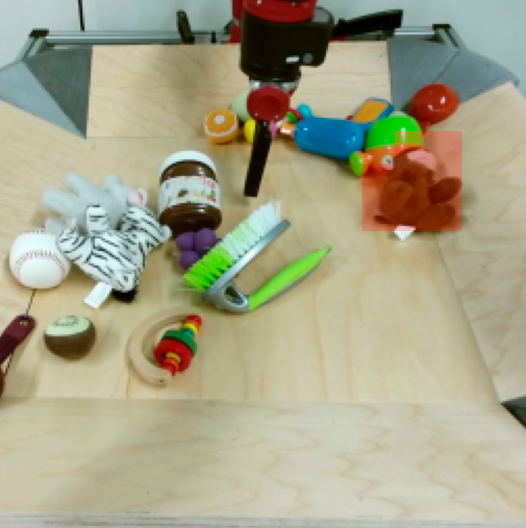} & &
     \includegraphics[width=\curWidth]{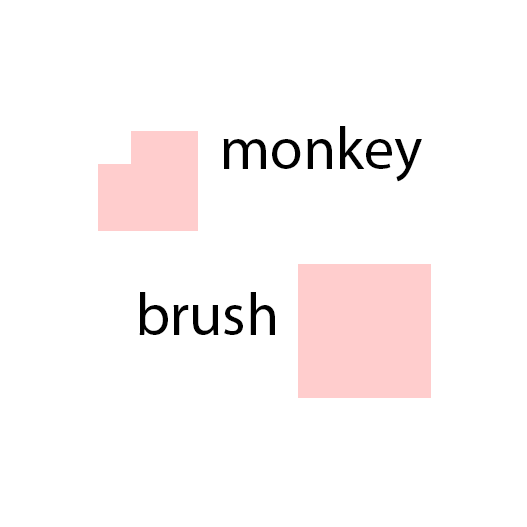} & 
     
     \includegraphics[width=\curWidth]{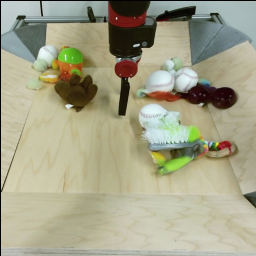} & 
     \includegraphics[width=\curWidth]{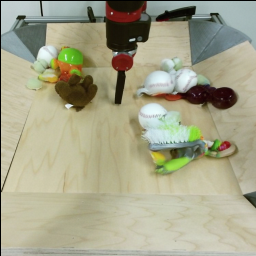} & 
     \includegraphics[width=\curWidth]{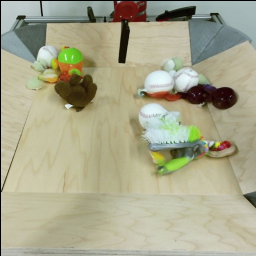} \\
     \includegraphics[width=\curWidth]{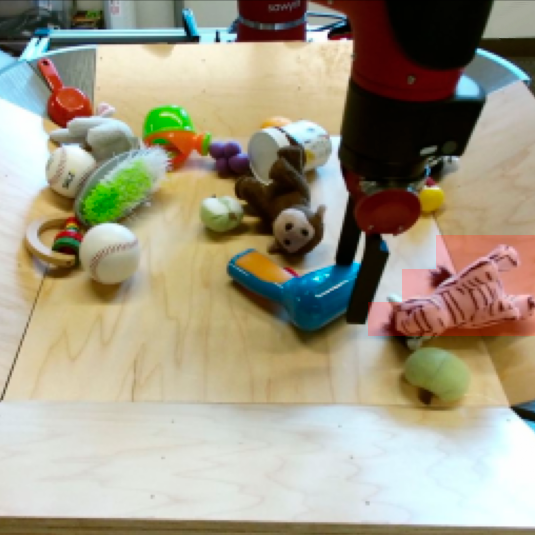} &
     \includegraphics[width=\curWidth]{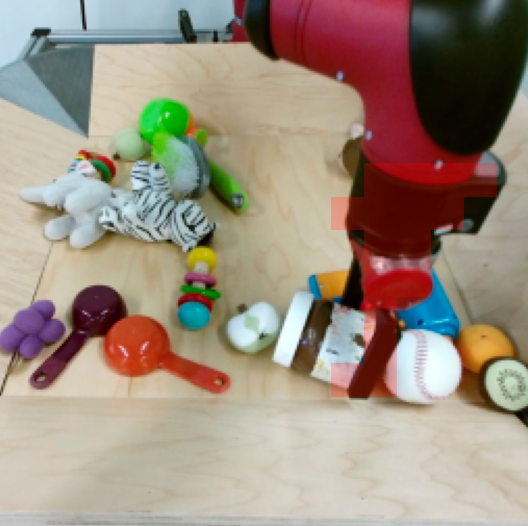} & &
     \includegraphics[width=\curWidth]{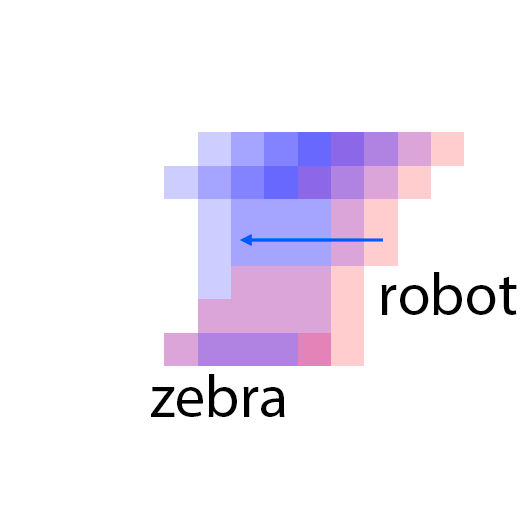} & 
     
     \includegraphics[width=\curWidth]{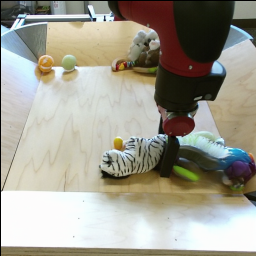} & 
     \includegraphics[width=\curWidth]{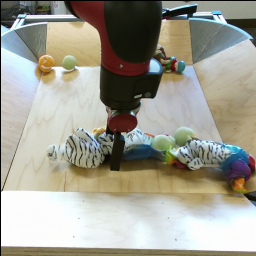} & 
     \includegraphics[width=\curWidth]{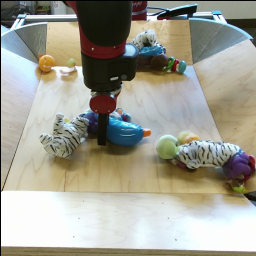} \\
     \includegraphics[width=\curWidth]{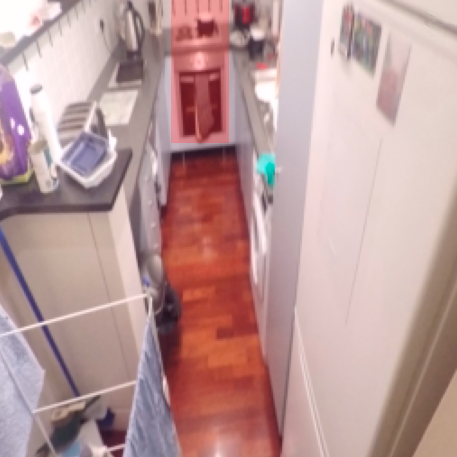} & & &
     \includegraphics[width=\curWidth]{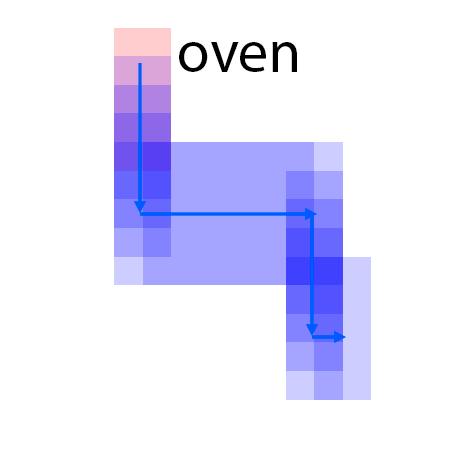} & 
     
     \includegraphics[width=\curWidth]{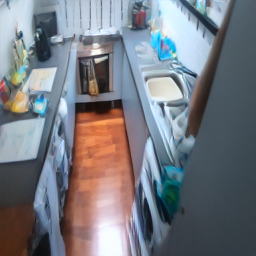} & 
     \includegraphics[width=\curWidth]{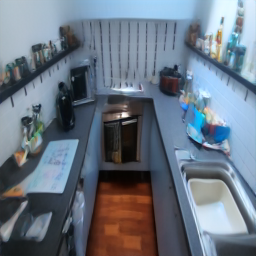} & 
     \includegraphics[width=\curWidth]{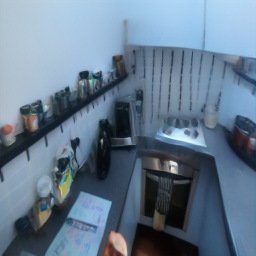} \\
     \includegraphics[width=\curWidth]{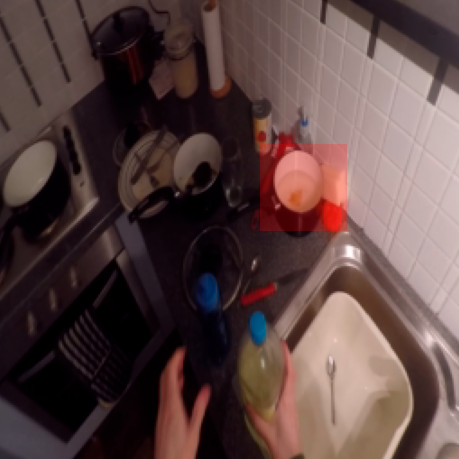} & & &
     \includegraphics[width=\curWidth]{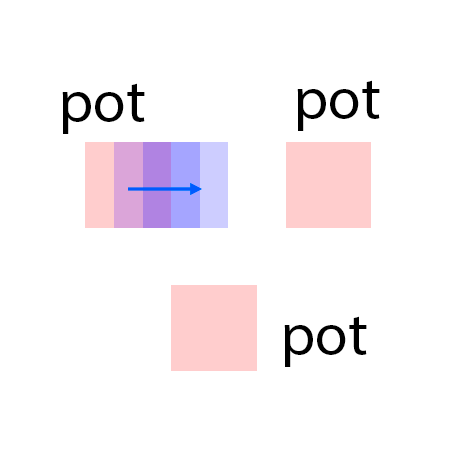} & 
     
     \includegraphics[width=\curWidth]{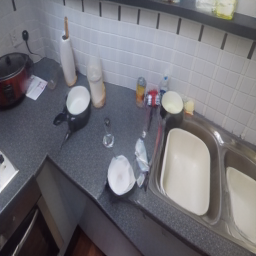} & 
     \includegraphics[width=\curWidth]{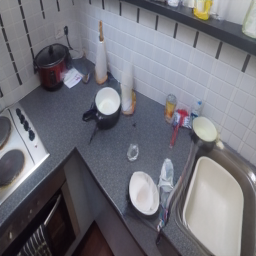} & 
     \includegraphics[width=\curWidth]{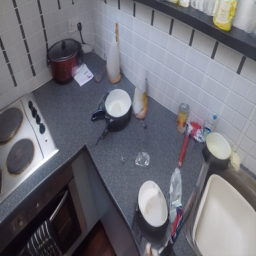} \\
\end{tabular}
\caption{Composition and animation with \methodName on different datasets. Red patches correspond to the copied features from the source images. Blue patches correspond to the intended future locations of the red patches. Blue arrows only illustrate the direction of the motion, but are not used as conditioning signals in the model.  
}\label{fig:composition_and_animation}
\end{figure*}

\noindent\textbf{Composition and Animation.} Fig.~\ref{fig:composition_and_animation} shows additional examples of videos generated with \methodName by composing scenes from object features adopted from other images. In Fig.~\ref{fig:epic} we show examples of videos generated under different control scenarios on the EPIC-KITCHENS dataset, including camera and hand motion.

\begin{figure*}[t]
    \centering
    \newcommand\curWidth{2cm}
    \begin{tabular}{@{}c@{\hspace{0.5mm}}c@{\hspace{0.5mm}}c@{\hspace{0.5mm}}c@{\hspace{0.5mm}}c@{\hspace{0.5mm}}c@{\hspace{0.5mm}}c@{\hspace{0.5mm}}c@{}}
        & \multicolumn{6}{c}{generated sequence$\rightarrow$} \\
         \includegraphics[width=\curWidth]{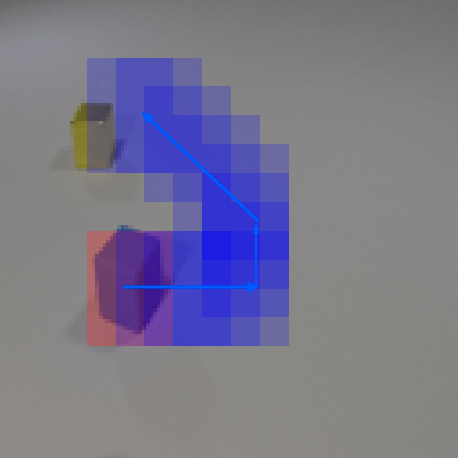} &
         \includegraphics[width=\curWidth]{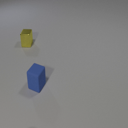} &
         \includegraphics[width=\curWidth]{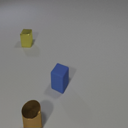} & \includegraphics[width=\curWidth]{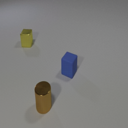} & \includegraphics[width=\curWidth]{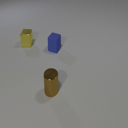} & \includegraphics[width=\curWidth]{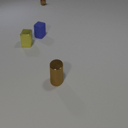} & \includegraphics[width=\curWidth]{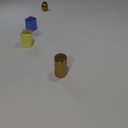} \\
         \includegraphics[width=\curWidth]{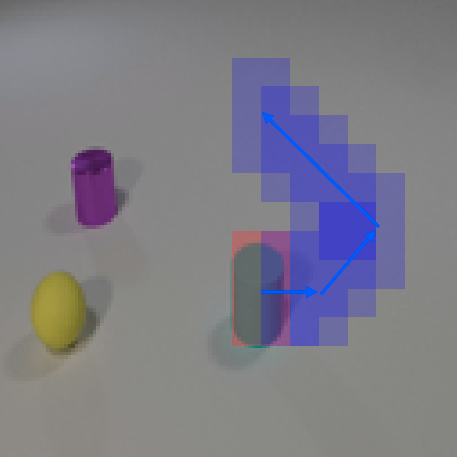} &
         \includegraphics[width=\curWidth]{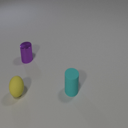} &
         \includegraphics[width=\curWidth]{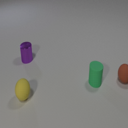} & \includegraphics[width=\curWidth]{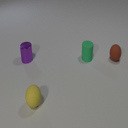} & \includegraphics[width=\curWidth]{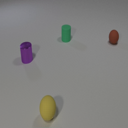} & \includegraphics[width=\curWidth]{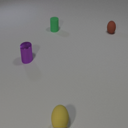} & \includegraphics[width=\curWidth]{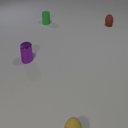} \\
         \includegraphics[width=\curWidth]{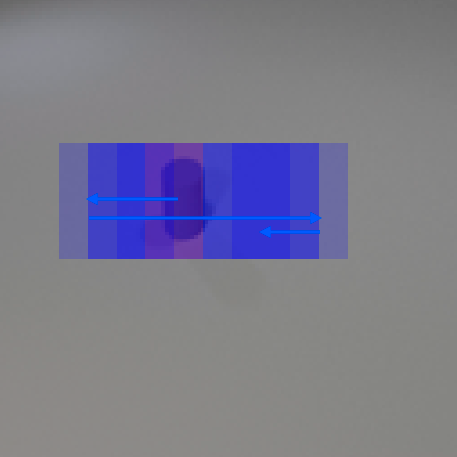} &
         \includegraphics[width=\curWidth]{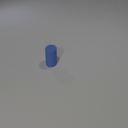} &
         \includegraphics[width=\curWidth]{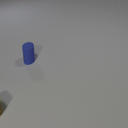} & \includegraphics[width=\curWidth]{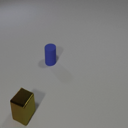} & \includegraphics[width=\curWidth]{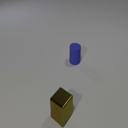} & \includegraphics[width=\curWidth]{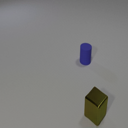} & \includegraphics[width=\curWidth]{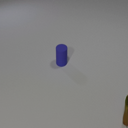} \\
         \includegraphics[width=\curWidth]{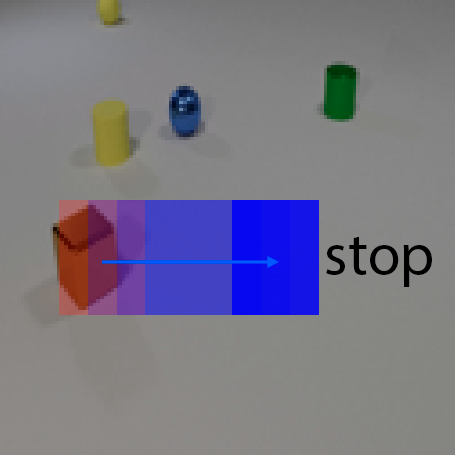} &
         \includegraphics[width=\curWidth]{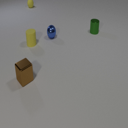} &
         \includegraphics[width=\curWidth]{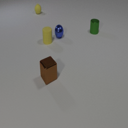} & \includegraphics[width=\curWidth]{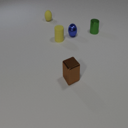} & \includegraphics[width=\curWidth]{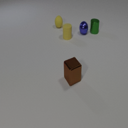} & \includegraphics[width=\curWidth]{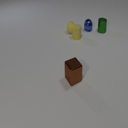} & \includegraphics[width=\curWidth]{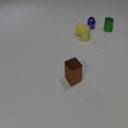} \\

    \end{tabular}
    \caption{Generalization to o.o.d. controls on the CLEVRER~\cite{clevrer} dataset. 
    }
    \label{fig:ood_clevrer}
\end{figure*}

\noindent\textbf{Out of Distribution Controls.} In the main paper we show the generalization capabilities of \methodName to o.o.d. controls in the BAIR~\cite{Ebert2017SelfSupervisedVP} dataset, such as moving background objects. Here we additionally provide videos generated from o.o.d. controls in the CLEVRER~\cite{clevrer} dataset (see Fig.~\ref{fig:ood_clevrer}). These controls include curved trajectories and sudden direction and velocity changes (\eg moving back and stopping).

\begin{figure*}[t]
    \centering
    \newcommand\curWidth{2.5cm}
    \begin{tabular}{@{}r@{\hspace{0.5mm}}c@{\hspace{0.5mm}}c@{\hspace{0.5mm}}c@{\hspace{0.5mm}}c@{\hspace{0.5mm}}c@{\hspace{0.5mm}}c@{\hspace{0.5mm}}c@{}}
        & \multicolumn{6}{c}{generated sequence$\rightarrow$} \\
         \rotatebox{90}{\makebox[\curWidth][c]{\makecell[c]{\methodName (ours)}}} &
         \includegraphics[width=\curWidth]{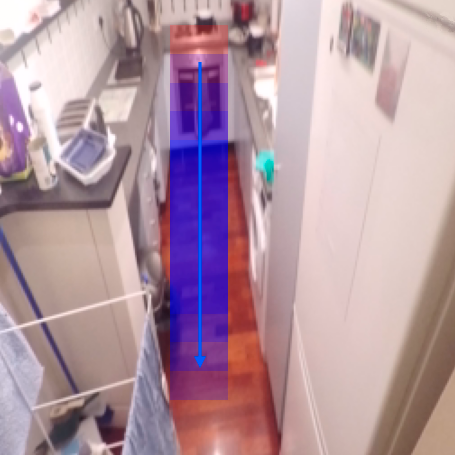} & 
         \includegraphics[width=\curWidth]{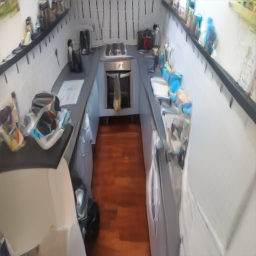} & 
         \includegraphics[width=\curWidth]{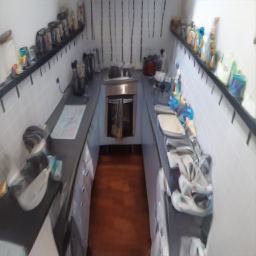} & \includegraphics[width=\curWidth]{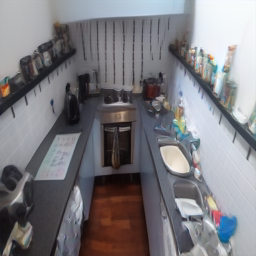} & \includegraphics[width=\curWidth]{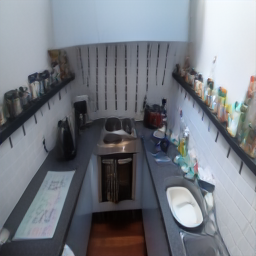} & \includegraphics[width=\curWidth]{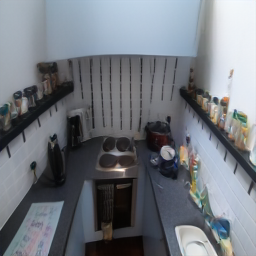} \\
         \rotatebox{90}{\makebox[\curWidth][c]{\makecell[c]{YODA}}} &
         \includegraphics[width=\curWidth]{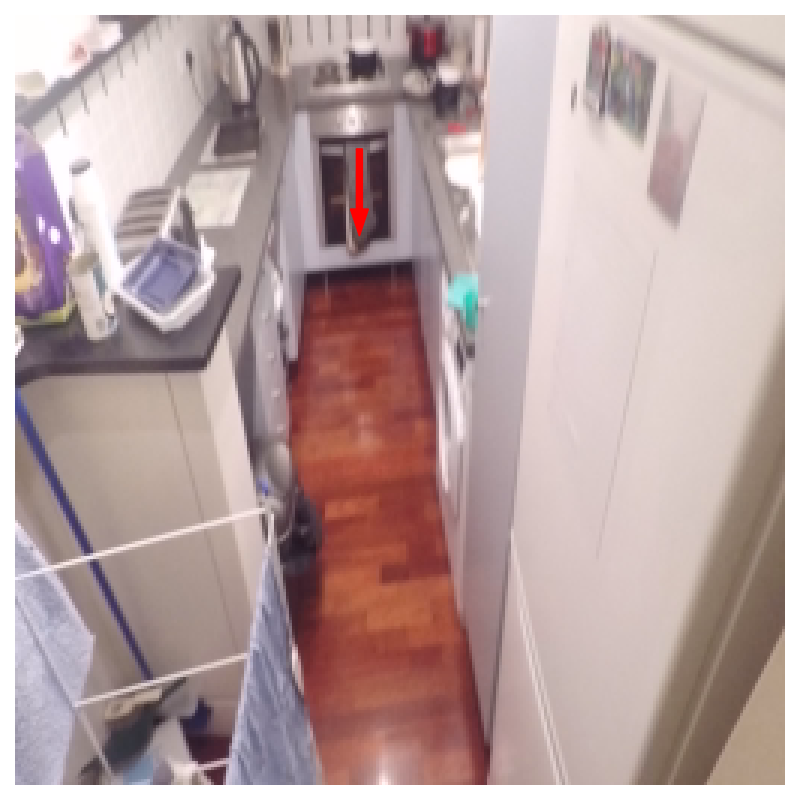} & 
         \includegraphics[width=\curWidth]{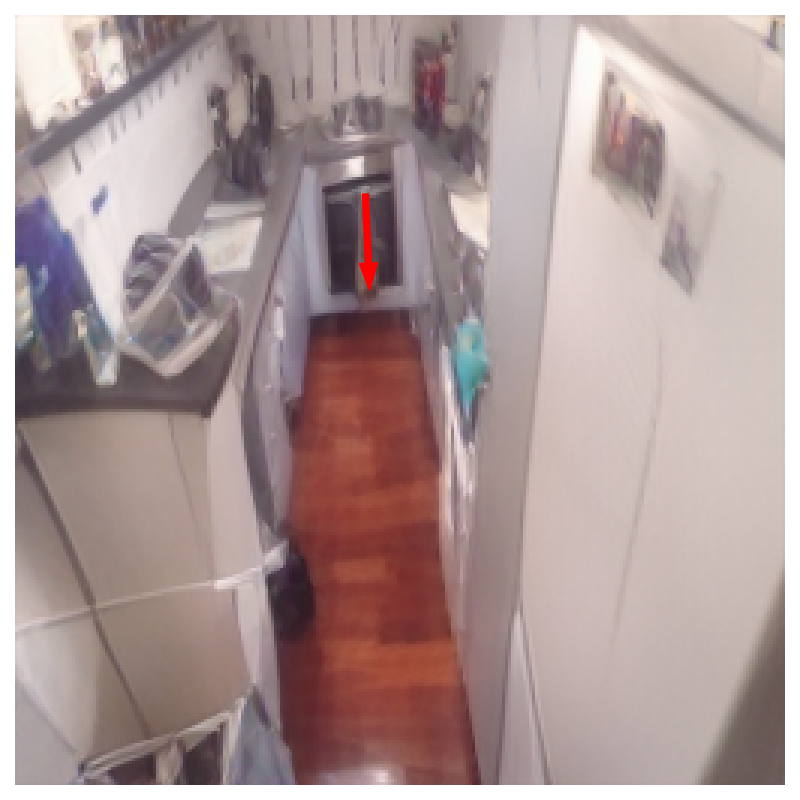} & 
         \includegraphics[width=\curWidth]{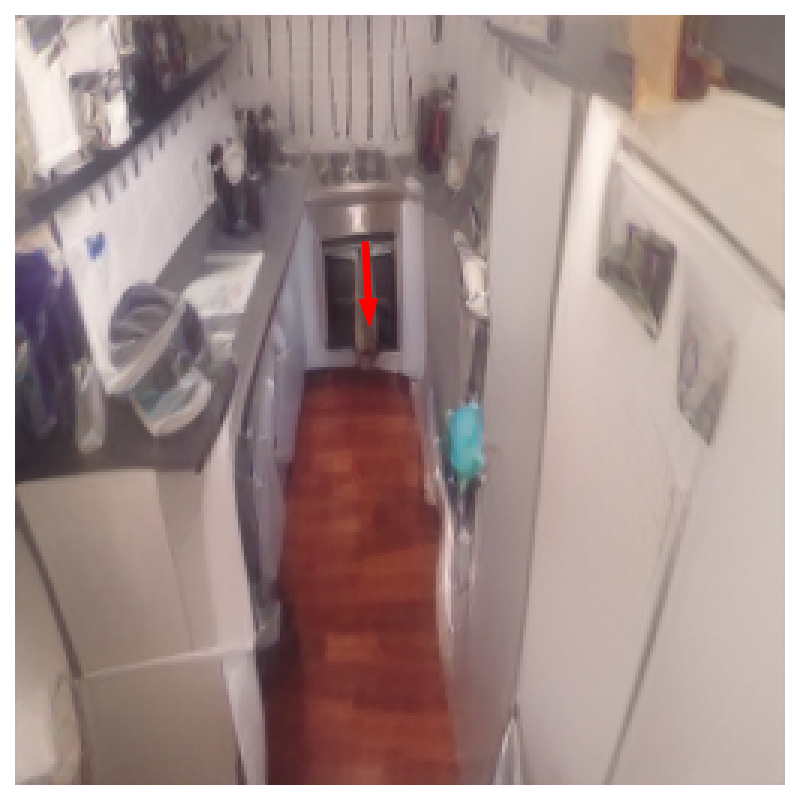} & \includegraphics[width=\curWidth]{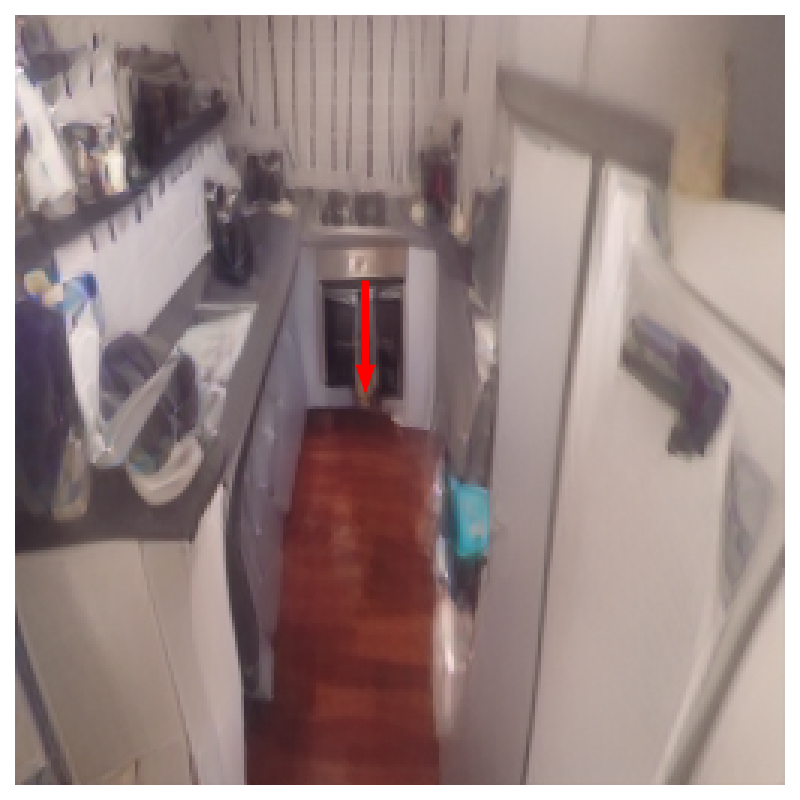} & \includegraphics[width=\curWidth]{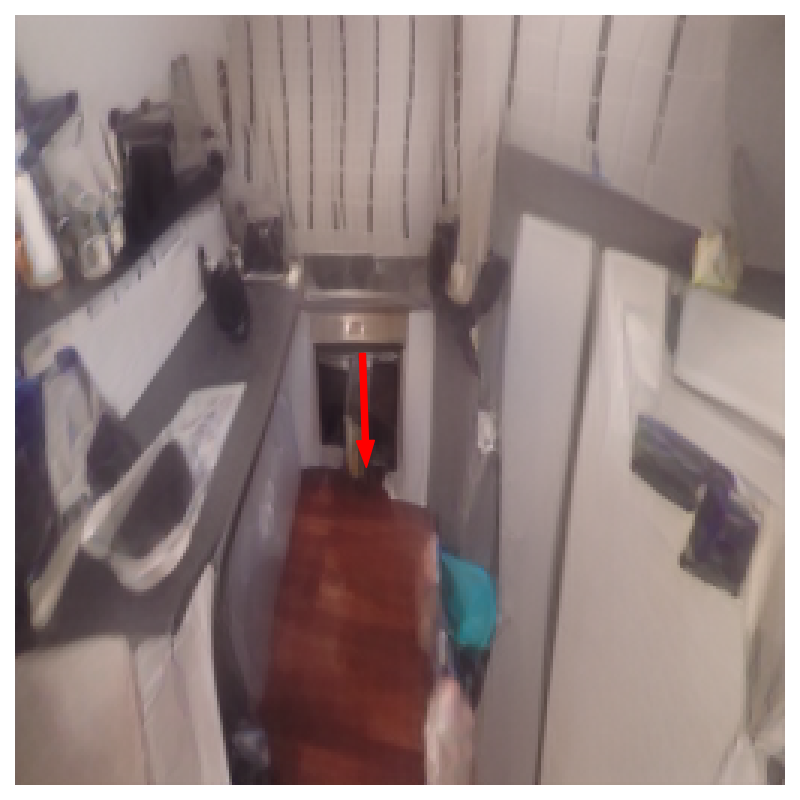} & \includegraphics[width=\curWidth]{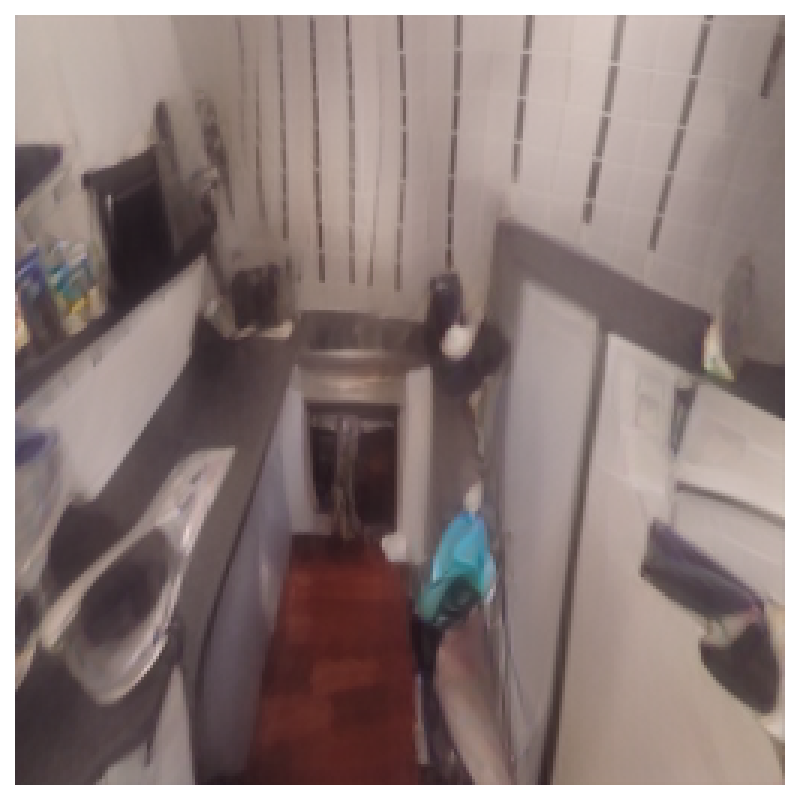} \\
         \rotatebox{90}{\makebox[\curWidth][c]{\makecell[c]{\methodName (ours)}}} &
         \includegraphics[width=\curWidth]{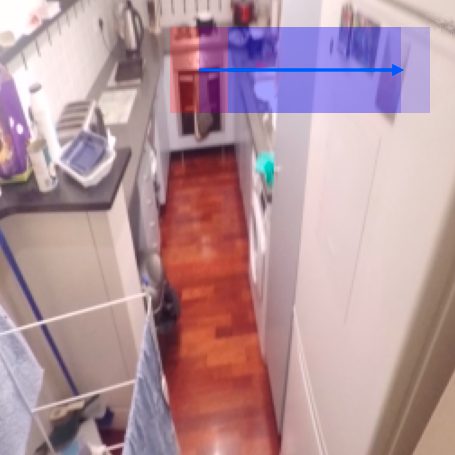} & 
         \includegraphics[width=\curWidth]{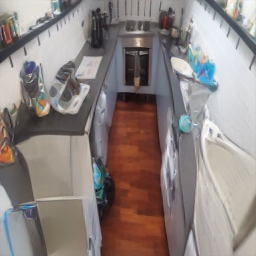} & 
         \includegraphics[width=\curWidth]{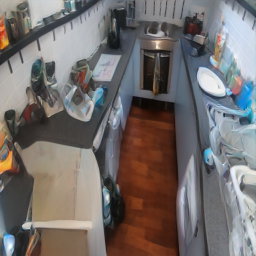} & \includegraphics[width=\curWidth]{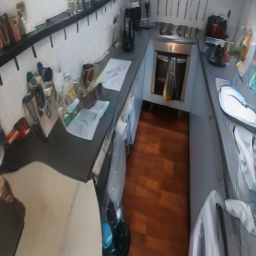} & \includegraphics[width=\curWidth]{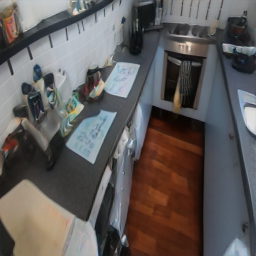} & \includegraphics[width=\curWidth]{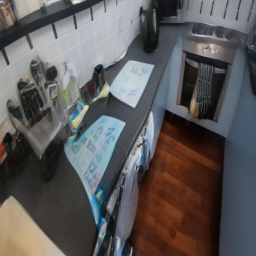} \\
         \rotatebox{90}{\makebox[\curWidth][c]{\makecell[c]{YODA}}} &
         \includegraphics[width=\curWidth]{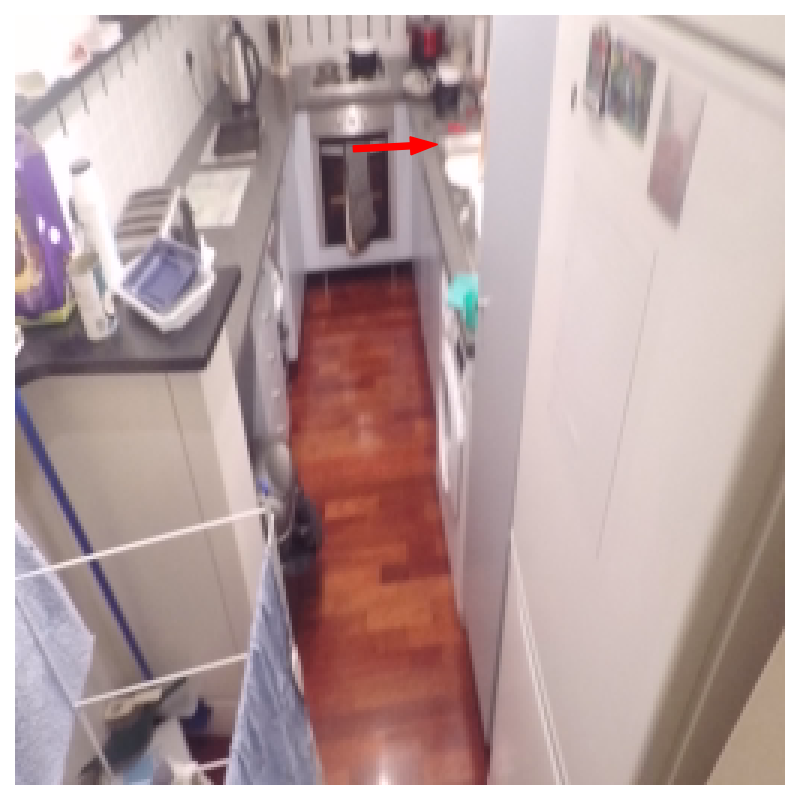} & 
         \includegraphics[width=\curWidth]{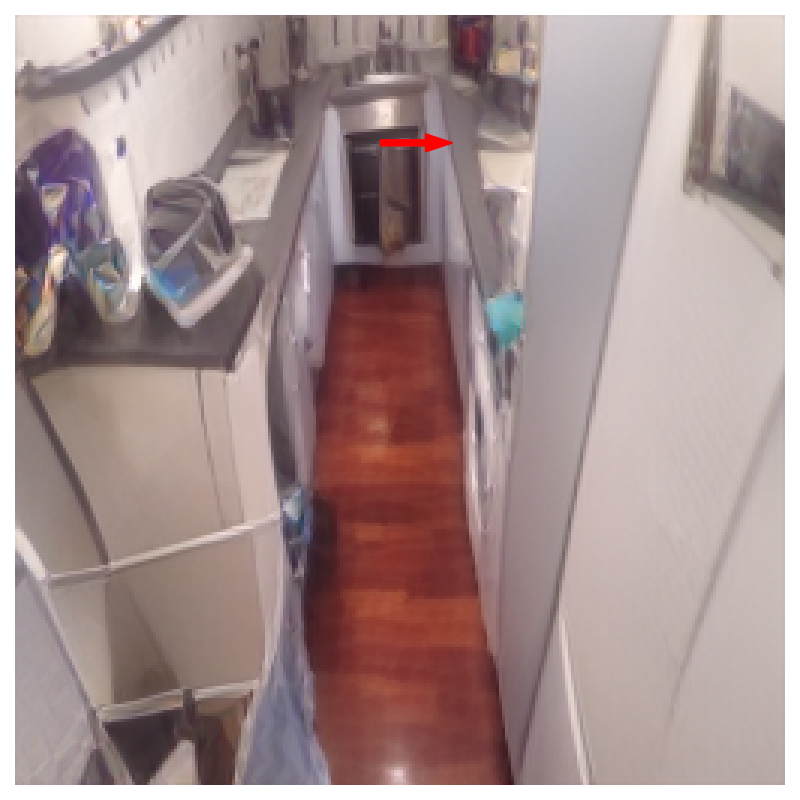} & 
         \includegraphics[width=\curWidth]{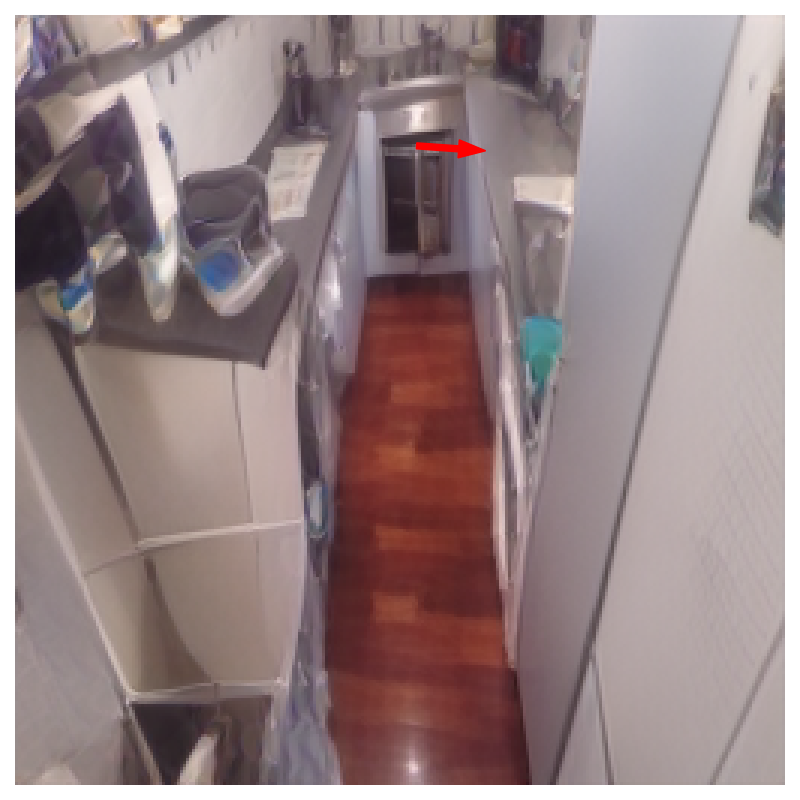} & \includegraphics[width=\curWidth]{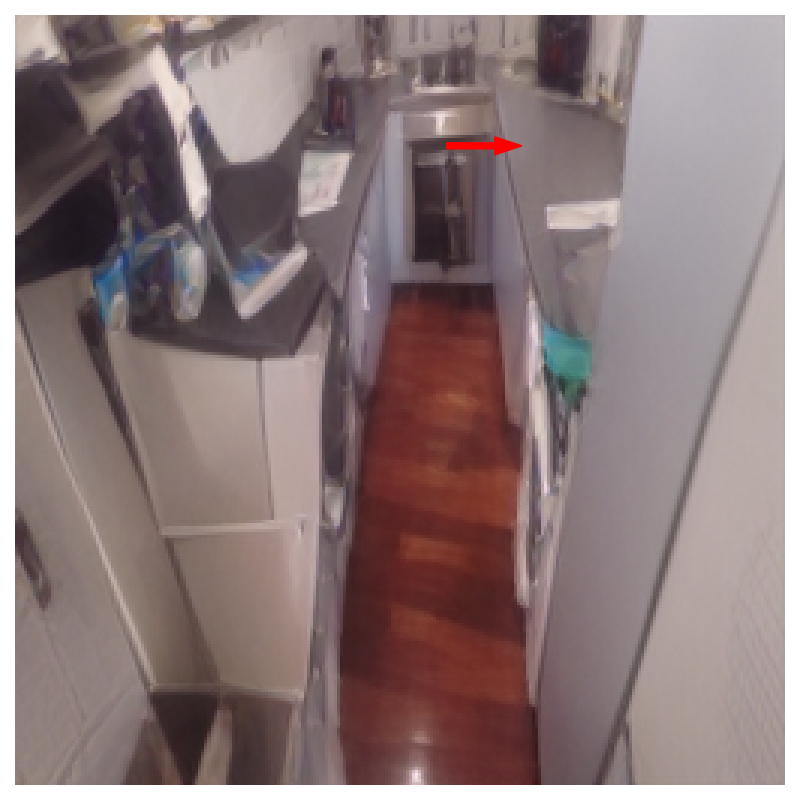} & \includegraphics[width=\curWidth]{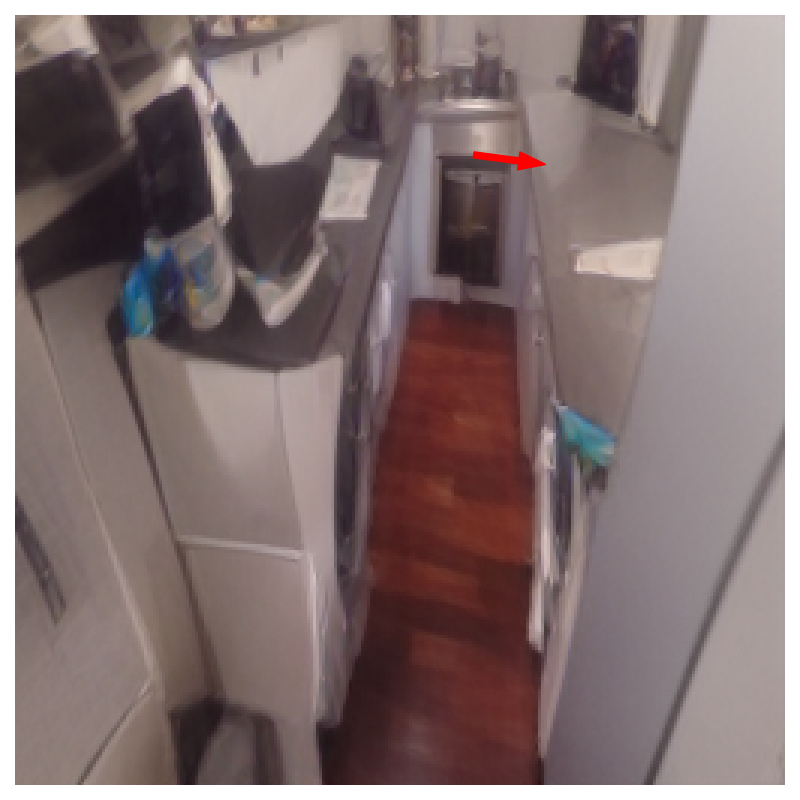} & \includegraphics[width=\curWidth]{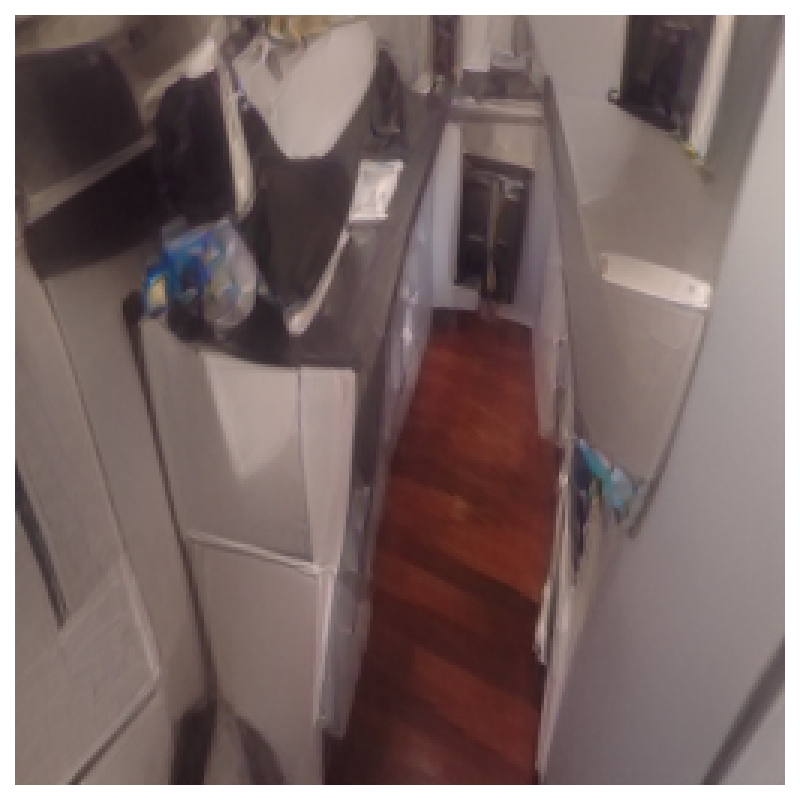} \\

    \end{tabular}
    \caption{Comparison to YODA~\cite{davtyan2024learn} on controllable video generation on the EPIC-KITCHENS~\cite{epic} dataset. \methodName generates more realistic sequences, while following the control. 
    }
    \label{fig:yoda_comparison}
\end{figure*}

\noindent\textbf{Comparison to Prior Work.} In the main paper we provided some quantitative comparisons to prior work. Here we accomplish this by visually comparing the videos generated with our method to those generated with YODA~\cite{davtyan2024learn} (see Fig.~\ref{fig:yoda_comparison}) on the EPIC-KITCHENS~\cite{epic} dataset. Notice that our model generates more temporally-consistent and realistic videos, while accurately following the control.

\begin{figure*}[t]
    \centering
    \newcommand\curWidth{2cm}
    \begin{tabular}{@{}c@{\hspace{0.5mm}}c@{\hspace{0.5mm}}c@{\hspace{0.5mm}}c@{\hspace{0.5mm}}c@{\hspace{0.5mm}}c@{\hspace{0.5mm}}c@{}}
        source image & control & \multicolumn{5}{c}{generated sequence$\rightarrow$} \\
         \includegraphics[height=\curWidth, trim=0.0cm 0cm 0.0cm 0cm, clip]{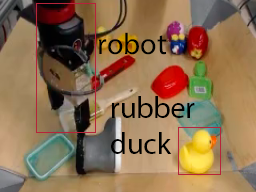} & \includegraphics[width=\curWidth]{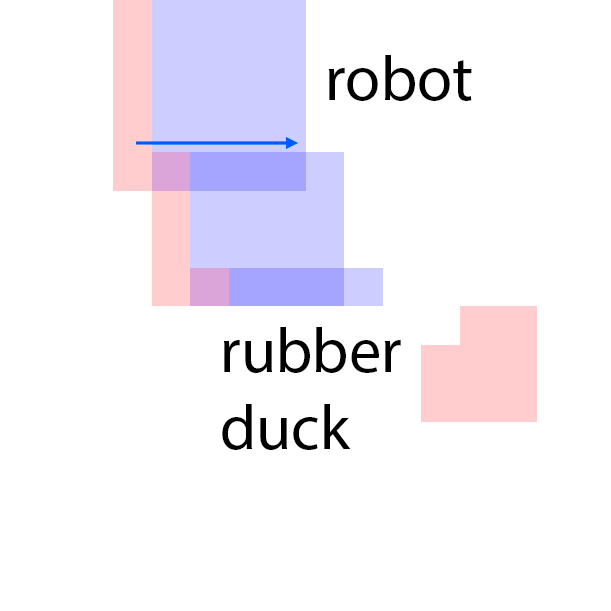} & 
         \includegraphics[width=\curWidth]{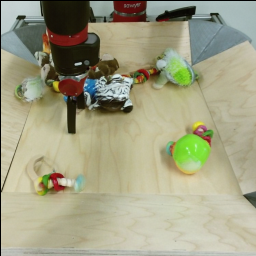} & 
         \includegraphics[width=\curWidth]{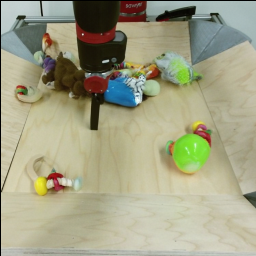} & 
         \includegraphics[width=\curWidth]{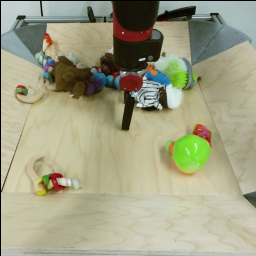} & 
         \includegraphics[width=\curWidth]{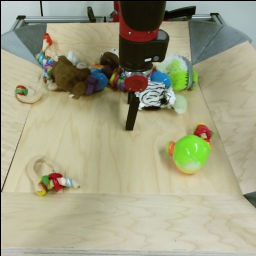} & 
         \includegraphics[width=\curWidth]{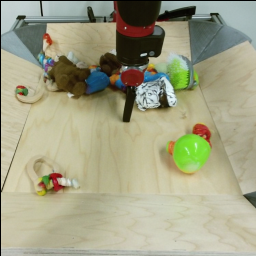} \\
         \includegraphics[height=\curWidth, trim=0.0cm 0cm 0.0cm 0cm, clip]{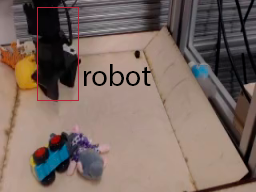} & \includegraphics[width=\curWidth]{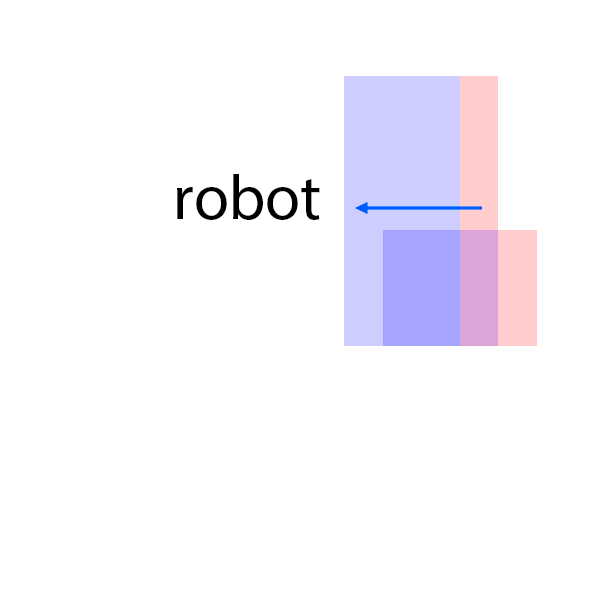} & 
         \includegraphics[width=\curWidth]{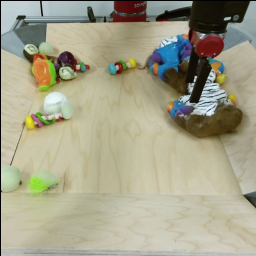} & 
         \includegraphics[width=\curWidth]{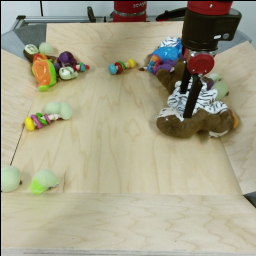} & 
         \includegraphics[width=\curWidth]{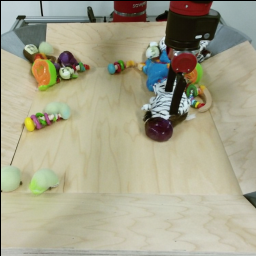} & 
         \includegraphics[width=\curWidth]{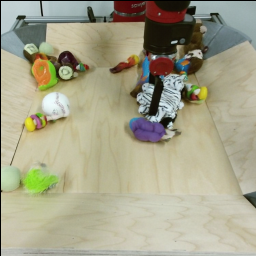} & 
         \includegraphics[width=\curWidth]{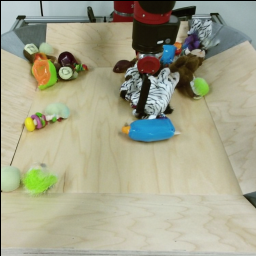} \\
         \includegraphics[height=\curWidth, trim=0.0cm 0cm 0.0cm 0cm, clip]{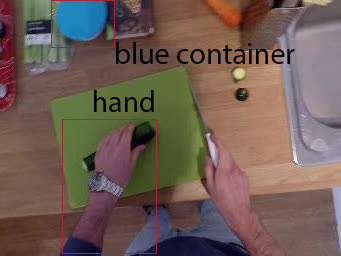} & \includegraphics[width=\curWidth]{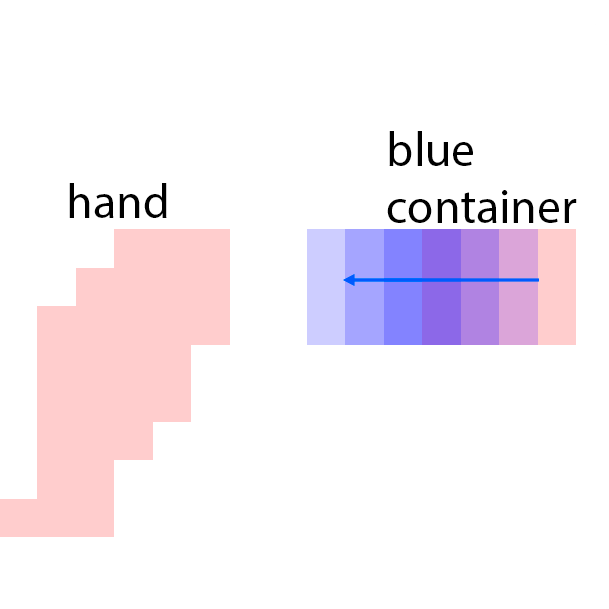} & 
         \includegraphics[width=\curWidth]{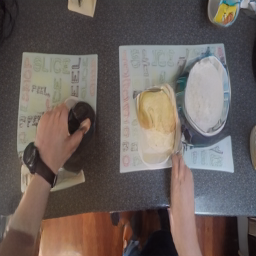} & 
         \includegraphics[width=\curWidth]{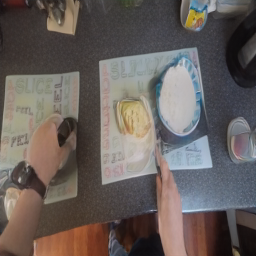} & 
         \includegraphics[width=\curWidth]{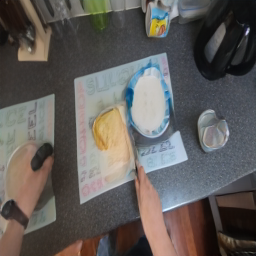} & 
         \includegraphics[width=\curWidth]{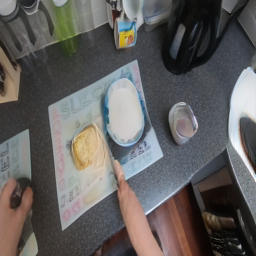} & 
         \includegraphics[width=\curWidth]{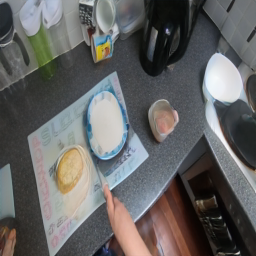} \\
         \includegraphics[height=\curWidth, trim=0.0cm 0cm 0.0cm 0cm, clip]{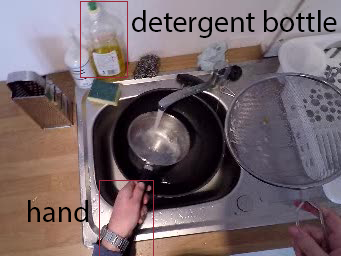} & \includegraphics[width=\curWidth]{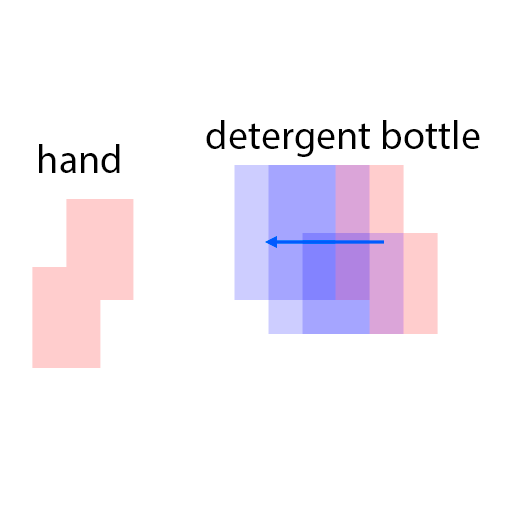} & 
         \includegraphics[width=\curWidth]{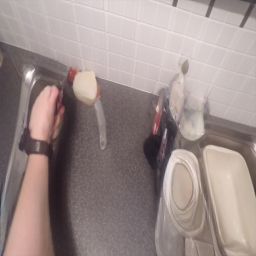} & 
         \includegraphics[width=\curWidth]{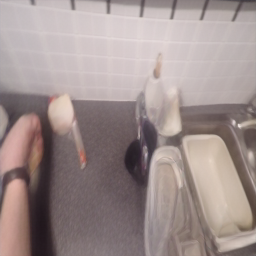} & 
         \includegraphics[width=\curWidth]{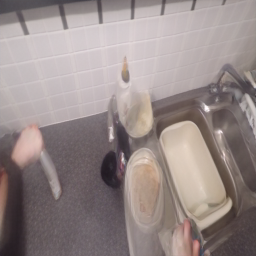} & 
         \includegraphics[width=\curWidth]{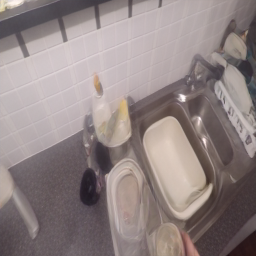} & 
         \includegraphics[width=\curWidth]{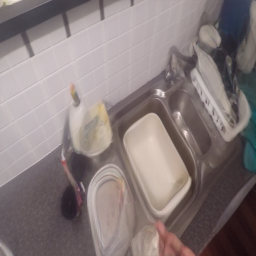} \\
         
    \end{tabular}
    \caption{Examples of cross-domain transfer on the BAIR~\cite{Ebert2017SelfSupervisedVP} and the EPIC-KITCHENS~\cite{epic} datasets. Notice how \methodName changes the appearances of the objects when transferring them to the familiar domain (\eg the color of the watch on the hand). 
    }
    \label{fig:cross_new}
\end{figure*}

\noindent\textbf{Cross-domain Transfer.} Thanks to the use of DINOv2~\cite{Oquab2023DINOv2LR} features for conditioning, at inference our model is able to compose scenes from features adopted from out of domain images in a zero-shot manner. In the main paper we provided examples of this emerging capability of \methodName on the CLEVRER~\cite{clevrer} dataset. Here we demonstrate that this capability also extends to more complex data, such as transferring from other robotic scenes from the ROBONET~\cite{dasari2019robonet} dataset to BAIR, or using features from other kitchens in the EPIC-KITCHENS~\cite{epic} dataset to compose and animate scenes in the kitchen the model was trained on (see Fig.~\ref{fig:cross_new}).

\noindent\textbf{Long Video Generation.} To demonstrate controllability of our method we opted for generating short videos, so that one is able to follow the objects and compare their motion to the control signal. However, this might have caused an impression that our model is only able to generate short videos. In order to address this issue, we provide some extremely long (512 frames) videos generated with \methodName on the CLEVRER~\cite{clevrer} dataset. In order not to overload the PDF, we only provide the videos via the webpage (check the ``Long Video Generation'' section on the website). Despite the fact that \methodName was only trained on short (3 frames) windows, long video generation is possible with our method, as it correlates subsequent windows of frames via conditioning on the reference and the context frames. Thus, \methodName can qualify as an autoregressive model.

\begin{figure*}[t]
    \centering
    \newcommand\curWidth{2cm}
    \begin{tabular}{@{}c@{\hspace{0.5mm}}c@{\hspace{0.5mm}}c@{\hspace{0.5mm}}c@{\hspace{0.5mm}}c@{\hspace{0.5mm}}c@{\hspace{0.5mm}}c@{}}
        source image & control & \multicolumn{5}{c}{generated sequence$\rightarrow$} \\
         \includegraphics[height=\curWidth, trim=0.0cm 0cm 0.0cm 0cm, clip]{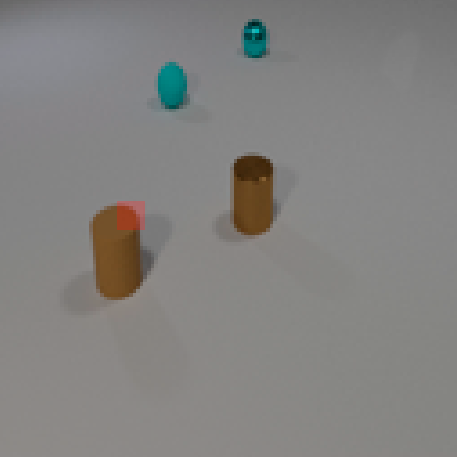} & \includegraphics[width=\curWidth]{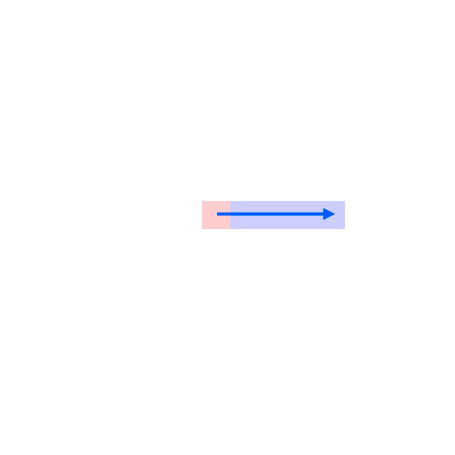} & 
         \includegraphics[width=\curWidth]{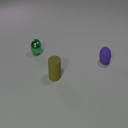} & 
         \includegraphics[width=\curWidth]{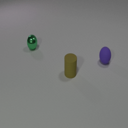} & 
         \includegraphics[width=\curWidth]{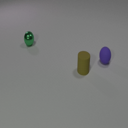} & 
         \includegraphics[width=\curWidth]{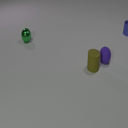} & 
         \includegraphics[width=\curWidth]{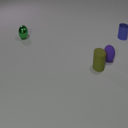} \\
         \includegraphics[height=\curWidth, trim=0.0cm 0cm 0.0cm 0cm, clip]{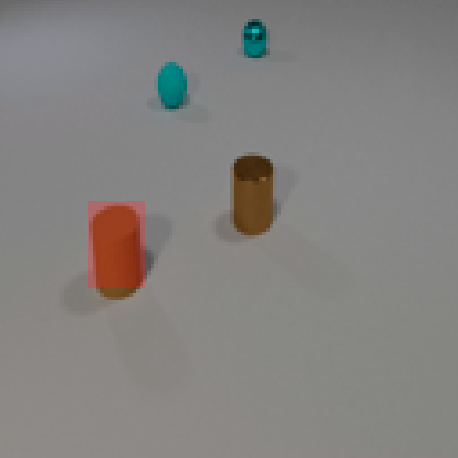} & \includegraphics[width=\curWidth]{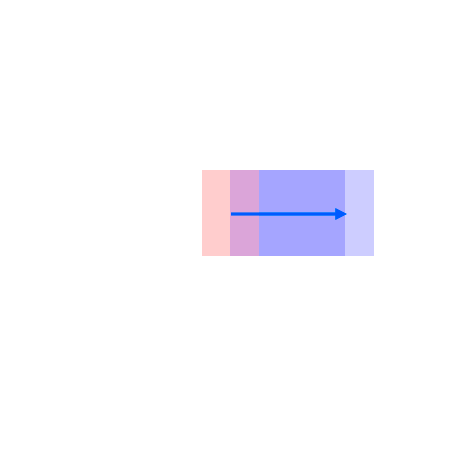} & 
         \includegraphics[width=\curWidth]{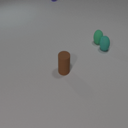} & 
         \includegraphics[width=\curWidth]{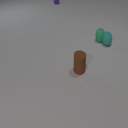} & 
         \includegraphics[width=\curWidth]{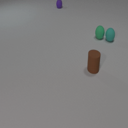} & 
         \includegraphics[width=\curWidth]{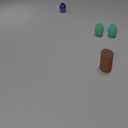} & 
         \includegraphics[width=\curWidth]{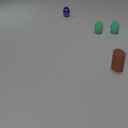} \\
         \includegraphics[height=\curWidth, trim=0.0cm 0cm 0.0cm 0cm, clip]{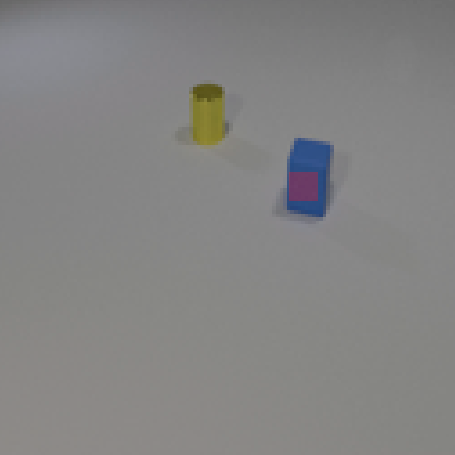} & \includegraphics[width=\curWidth]{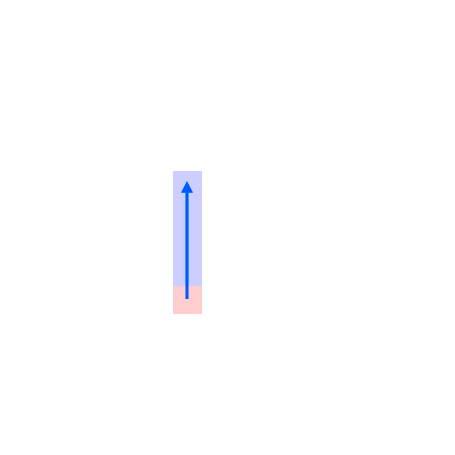} & 
         \includegraphics[width=\curWidth]{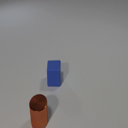} & 
         \includegraphics[width=\curWidth]{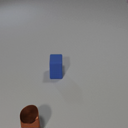} & 
         \includegraphics[width=\curWidth]{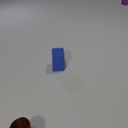} & 
         \includegraphics[width=\curWidth]{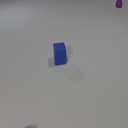} & 
         \includegraphics[width=\curWidth]{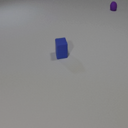} \\
         \includegraphics[height=\curWidth, trim=0.0cm 0cm 0.0cm 0cm, clip]{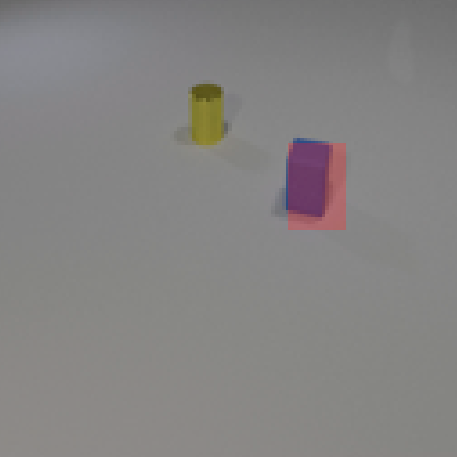} & \includegraphics[width=\curWidth]{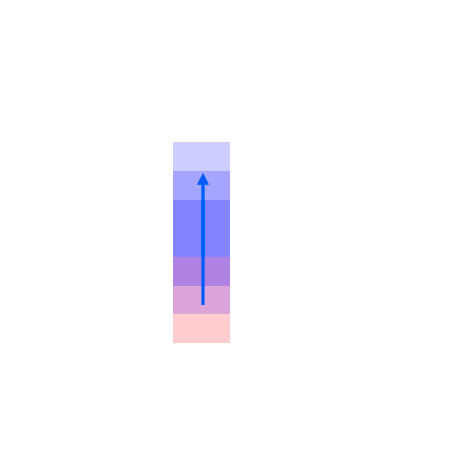} & 
         \includegraphics[width=\curWidth]{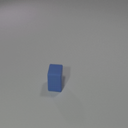} & 
         \includegraphics[width=\curWidth]{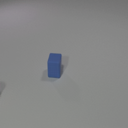} & 
         \includegraphics[width=\curWidth]{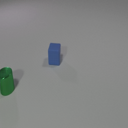} & 
         \includegraphics[width=\curWidth]{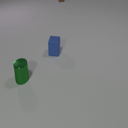} & 
         \includegraphics[width=\curWidth]{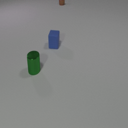} \\
         
    \end{tabular}
    \caption{The effect of the number of control tokens on the generated sequences. More tokens make the generated object more consistent with the source appearance (\eg in color and pose). 
    }
    \label{fig:nc_robustness}
\end{figure*}

\noindent\textbf{Robustness to the Number of Controls.} In Fig.~\ref{fig:nc_robustness} we show some generated sequences with only one vs many DINOv2~\cite{Oquab2023DINOv2LR} tokens provided per object. While \methodName is pretty robust to the number of controls provided and can inpaint the missing information, the sequences generated conditioned on more tokens are more consistent with the source image (\eg in the last video the blue cube does not rotate with more tokens provided, \ie the pose of the object is fixed given the features).

\bibliography{references}

\end{document}